\documentclass[letterpaper]{article} % DO NOT CHANGE THIS
\usepackage{aaai25}  % DO NOT CHANGE THIS
\usepackage{times}  % DO NOT CHANGE THIS
\usepackage{helvet}  % DO NOT CHANGE THIS
\usepackage{courier}  % DO NOT CHANGE THIS
\usepackage[hyphens]{url}  % DO NOT CHANGE THIS
\usepackage{graphicx} % DO NOT CHANGE THIS
\usepackage{natbib}  % DO NOT CHANGE THIS AND DO NOT ADD ANY OPTIONS TO IT
\usepackage{caption} % DO NOT CHANGE THIS AND DO NOT ADD ANY OPTIONS TO IT
\usepackage{algorithm}
\usepackage{algorithmic}

\usepackage{newfloat}
\usepackage{listings}
\usepackage{outlines}
\DeclareCaptionStyle{ruled}{labelfont=normalfont,labelsep=colon,strut=off} % DO NOT CHANGE THIS
\floatstyle{ruled}
\newfloat{listing}{tb}{lst}{}
\floatname{listing}{Listing}
\usepackage{url}
\usepackage[utf8]{inputenc}
\usepackage{adjustbox}
\usepackage{amsmath, bm} % assumes amsmath package installed
\usepackage{amssymb}  % assumes amsmath package installed
\usepackage{upgreek}
\usepackage{booktabs}
\usepackage{subcaption}
\usepackage{lipsum}
\usepackage[utf8]{inputenc}
\newtheorem{theorem}{Theorem}%[section]
\usepackage{svg}

\newtheorem{definition}{Definition}%[section]

\title{Learning Physics Informed Neural ODEs\\ With Partial Measurements}
\author {
    Paul Ghanem\textsuperscript{\rm 1},
    Ahmet Demirkaya\textsuperscript{\rm 1},
    Tales Imbiriba\textsuperscript{\rm 2},
    Alireza Ramezani\textsuperscript{\rm 2},
    Zachary Danziger\textsuperscript{\rm 3},
    Deniz Erdogmus\textsuperscript{\rm 2},
}
\affiliations {
    \textsuperscript{\rm 1}Northeastern University, Boston Massachusetts\\
    \textsuperscript{\rm 2}University of Massachusetts Boston, Boston Massachusetts \\
    \textsuperscript{\rm 3}Emory University, Atlanta Georgia\\    
     \{ghanem.p, demirkaya.a\}@northeastern.edu,
     Tales.Imbiriba@umb.edu,
     a.ramezani@northeastern.edu,
     zdanzige@fiu.edu,
     d.erdogmus@northeastern.edu
     }

\usepackage{bibentry}
\begin{document}

\maketitle

\begin{abstract}
Learning dynamics governing physical and spatiotemporal processes is a challenging problem, especially in scenarios where states are partially measured. In this work, we tackle the problem of learning dynamics governing these systems when parts of the system's states are not measured, specifically when the dynamics generating the non-measured states are unknown. Inspired by state estimation theory and Physics Informed Neural ODEs, we present a sequential optimization framework in which dynamics governing unmeasured processes can be learned. We demonstrate the performance of the proposed approach leveraging  numerical simulations and a real dataset extracted from an electro-mechanical positioning system. We show how the underlying equations fit into our formalism and demonstrate the improved performance of the proposed method when compared with baselines.
\end{abstract}


\section{Introduction}
% \vspace{-0.3cm}
Ordinary differential equations (ODEs) are used to describe the state evolution of many complex physical systems in engineering, biology, and other fields of natural sciences. 
Traditionally, first-principle notions are leveraged in designing ODEs as a form to impose physical meaning and interpretability~\citep{psichogios1992hybrid} of latent states. 
A major issue, however, is the inherent complexity of real-world problems for which even carefully designed ODE systems cannot account for all aspects of the true underlying physical phenomenon \citep{karniadakis2021physics}. Moreover, we often require prediction of systems whose dynamics are not fully understood or are partially unknown.

In this context, Neural ODEs
(NODEs)~\citep{chen2018neural} emerged as a powerful tool for learning complex correlations directly from the data, where residual neural networks (NNs) are used to parameterize the hidden ODEs' states. Extensions of NODE were developed to improve learning speed \citep{xia2021heavy,massaroli2021differentiable} and learning longtime dependencies in irregularly sampled time series\citep{xia2021heavy}.
A major challenge in learning NODEs arises when latent states of interest contribute indirectly to the measurements. This is the case when an unmeasured state influences a measured state. In this scenario, NODE's standard solutions, which are optimized using the adjoint method~\citep{boltyanskiy1962mathematical}, are compromised.  
Furthermore, NODE systems may have infinitely many solutions since parameters and unmeasured states are estimated jointly. 
As a consequence, even when the model is capable of fitting the data, unmeasured states cannot be accurately inferred without constraining the solution space~\citep{Demirkaya_retinaembc2021}.
To constrain the solution space, hybrid and physics informed neural ODEs were developed to incorporate physical knowledge of the system being learned when available\citep{sholokhov2023physics,o2022stochastic}. Despite their recent success in neural ODEs and neural networks in general, these methods lack the ability to learn dynamical systems under the partial measurements scenario when dynamics generating unmeasured states are unknown.
Moreover, hybrid and physics informed strategies were leveraged to obtain estimations of missing states under partial measurements scenario~\citep{Imbiriba_hybrid_2022, Demirkaya_retinaembc2021,ghanem2021efficient}.
Despite the lack of a clear formalization, in these works the authors were imposing some kind of identifiability among states by adding known parts of the dynamics, resulting in hybrid first-principle data-driven models. Nevertheless, these works focus on state estimation using data-driven components to improve or augment existing dynamics but fail to learn global models and do not scale well for large models.

In this paper, we propose a sequential optimization approach that at each time step solves an alternating optimization problem for learning system dynamics under partially measured states, when states are identifiable.  The approach focuses on learning unknown dynamics from data where the state related to the unknown dynamics is unmeasured. Since the unobserved dynamics are unknown, we assume it is described by parametric models such as NNs. We propose aiding a model’s training with the knowledge of the physics regarding the measured states using a physics-informed loss term. The proposed solution leverages the relationship between many recursive state-space estimation procedures and Newton's method~\citep{humpherys2012fresh} to develop an efficient recursive NODE learning approach capable of sequentially learning states and model parameters. The benefit of the sequential strategy is twofold: 
$(1)$ reduce the need for accurate initial conditions during training;
$(2)$ enables usage of hidden states estimates to learn model parameters instead of simultaneous estimation of states and parameters, making second-order optimization methods feasible \textcolor{black}{under partial measurement scenario}.  
% \sout{Differently from standard recursive state estimation strategies \citep{humpherys2012fresh} where all states are jointly optimized,}
Furthermore, the proposed approach exploits the identifiable property of states by designing an alternating optimization strategy with respect to states and parameters.
%, where optimization updates performed over states precede parameter updates. 
%
% Unlike \citep{humpherys2012fresh} where optimization variables are combined and optimized jointly, our procedure optimizes latent states  assuming they are distinguishable using provided training data and initial/previous model parameters, then uses recovered latent state information to optimize model parameters. 
The result is an interconnected sequential optimization procedure, where at each step model parameters and data are used to estimate latent states, and corrected latent states are used to update the model parameters in the current optimization step. 
%
% Such alternating optimization approach aids the optimization of system parameters since it estimates unmeasured hidden states and uses them in learning system parameters. In the case of our proposed mechanism, it also prevents vanishing gradients.
% The alternating optimization aids the learning of the system's parameters since estimation of unmeasured hidden states can be used. 
% The proposed mechanism also prevents vanishing gradients.
Moreover, we define identifiable latent variables and test our proposed approach in hybrid scenarios where NNs replace parts of the ODE systems such that the identifiability of latent variables is kept. 
Finally, as a side effect of the recursive paradigm adopted the proposed strategy can assimilate data and estimate initial conditions by leveraging its sequential state estimation framework over past data. 
%\vspace{-0.5cm}
%\vspace{-0.2cm}
\section{Related Work}
%\vspace{-0.4cm}
\textbf{Partial Measurements:}
%\vspace{-0.3cm}
In the context of data-driven ODE designs, most learning frameworks assume that all states are measured in the sense that they are directly measured. This assumption does not reflect many real-world scenarios where a subset of the states are unmeasured. 
GP-SSM is a well-established approach used for dynamic systems identification \citep{mchutchon2015nonlinear,ialongo2019overcoming}. GP-SSM can be adapted by introducing a recognition model that maps outputs to latent states to solve the problem of partial measurements \citep{eleftheriadis2017identification}. Nevertheless, these methods do not scale well with large datasets and are limited to small trajectories\citep{doerr2018probabilistic}. Indeed, \citep{doerr2018probabilistic} minimizes this problem by using stochastic gradient ELBO optimization on minibatches. 
However,  
GP-SSM-based methods avoid learning the vector field describing the latent states and instead directly learn a mapping from a history of past inputs and measurements to the next measurement.
% i.e.  $x_{k+1}=f(x_k,u_k)$. \textcolor{black}{keep this or delete ?}

Similar approaches to recognition models have been used for Bayesian extensions of Neural Ordinary Differential Equations (NODEs). In these extensions, the NODE describes the dynamics of latent states, while the distribution of the initial latent variable given the measurements are approximated by encoder and decoder networks \citep{yildiz2019ode2vae,norcliffe2021neural}.
% Similar approaches to recognition models have been used for Bayesian extensions of NODEs, where the NODE describes the dynamics of latent states while the distribution of the initial latent variable given the measurements and vice versa are approximated by encoder and decoder networks \citep{yildiz2019ode2vae,norcliffe2021neural}. 
The encoder network, which links measurements to latent
states by a deterministic mapping or by approximating the conditional distribution, can also be a Recurrent
Neural Network (RNN) \citep{rubanova2019latent,kim2021inferring,de2019gru}, or an autoencoder \citep{bakarji2023discovering}.
Despite focusing on mapping measurements to latent states with neural networks and autoencoders, these works were not demonstrated to learn parameterized models under partial measurements. Moreover, this parameterized line of work of mapping measurement to latent states suffers from unidentifiability problem since several latent inputs could lead to the same measurement. 
Recently, sparse approaches such as \citep{bakarji2022discovering} merged encoder networks to identify a parsimonious transformation of the hidden dynamics of partially measured latent states. Moreover, Nonlinear Observers and recognition models were combined with NODEs to learn dynamic model parameters from partial measurements while enforcing physical knowledge in the latent space \citep{buisson2022recognition}. 
Differently from the aforementioned methods, in this work, we propose a recursive alternating approach that uses alternating Newton updates to optimize a quadratic cost function with respect to states and model parameters. Furthermore, the proposed strategy provides a systematic way to estimate initial conditions from historical data.

%\vspace{-0.2cm}
\textbf{Second order Newton method:}
Despite the efficiency and popularity of many stochastic gradient descent methods \citep{robbins1951stochastic,duchi2011adaptive,hinton2012neural,kingma2014adam} for optimizing NNs, great efforts have been devoted to exploiting second-order Newton methods where Hessian information is used, providing faster convergence \citep{martens2015optimizing,botev2017practical,gower2016stochastic,mokhtari2014res}. 
% Traditionally, stochastic gradient descent (SGD) \citep{robbins1951stochastic} and its variants are widely used for training neural networks. Among variants, we highlight such methods as AdaGrad \citep{duchi2011adaptive}, RMSprop \citep{hinton2012neural}, and Adam \citep{kingma2014adam}. Gradient descent methods are a special case of Newton's method where the Hessian is considered to be an Identity matrix(ref). Nonetheless, there has been a lot of effort to find ways to take advantage of second-order information in solving ML optimization problems \citep{martens2015optimizing,botev2017practical,gower2016stochastic,mokhtari2014res}. 
% Second-order Newton’s method is known to converge faster than first-order gradient based methods. 
When training neural networks, computing the inverse of the Hessian matrix can be extremely expensive \citep{goldfarb2020practical} or even intractable. To mitigate this issue, Quasi-Newton methods have been proposed to approximate the Hessian pre-conditioner matrix such as Shampoo algorithm~\citep{gupta2018shampoo}, which was extended in~\citep{anil2020scalable} to simplify blocks of the Hessian, and in  \citep{gupta2018shampoo} to be used in variational inference second-order approaches~\citep{peirson2022fishy}. Similarly, works in \citep{goldfarb2020practical,byrd2016stochastic} focused on developing stochastic quasi-Newton algorithms for problems with large amounts of data.  It was shown that recursive the extended Kalman filter can be viewed as Gauss-Newton method \citep{bell1994iterated,bertsekas1996incremental}. Moreover, Newton's method was used to derive recursive estimators for prediction and smoothing \citep{humpherys2012fresh}.
In this paper, we develop a recursive Newton method that mitigates the problem of partial measurements of latent states.

%\vspace{-0.3cm}
%\vspace{-0.2cm}
\section{Model and Background}
%\vspace{-0.3cm}

In this section, we describe our modeling assumptions, discuss the identifiability of latent states, and present the time evolution of the resulting generative model.
%\vspace{-0.15cm}
\subsection{Model}
%\vspace{-0.15cm}
In this work, we focus on dynamical models characterized by a set of ordinary differential equations describing the time evolution of system states $x(t)$ and system parameters  $\theta(t)$, and a measurement equation outputting measurements 
$y(t)\in\mathcal{Y}\subset\mathbb{R}^{d_y}$ of a subset of these states.  These models can be described as follows \citep{sarkka2023bayesian}:
%\vspace{-0.2cm}
\begin{equation}
    \begin{aligned}
        \dot{\theta}(t)&= \tilde{\nu}(t)\\[-10pt]
        \dot{x}(t)&=\underbrace{f(x(t),u(t),\overbrace{a(x(t),\theta(t))}^{\text{\color{red}Hidden physics}} )}_{\text{\color{blue}Known physics}} + \,\tilde{\epsilon}(t)\\[-3pt]
        y(t)&=h(x(t)) + \zeta(t)
    \label{eq:continuous_model}
    \end{aligned}
\end{equation}
where $x(t) \in \mathcal{X} \subset \mathbb{R}^{d_x}$ are systems states and $\theta(t)\in\mathcal{P} \subset \mathbb{R}^{d_\theta}$ are system parameters. 
$a:\mathcal{X} \times \mathcal{P} $ represents a system of hidden ODE parameterized by  $\theta(t)$ that needs to be learned without measurement available.
$f:\mathcal{X} \times \mathcal{P} \times \mathcal{U}$ represents a system of ODEs parameterized by  $\theta(t)$, where each equation describes the time dynamics of an individual component of a dynamical system.
$h: \mathcal{X}  \to \mathcal{Y}$ represents the measurement function. $u(t)\in\mathcal{U}\subset\mathbb{R}^{d_u}$ is a vector of external inputs, 
$\tilde{\nu}(t) \sim  \mathcal{N}(0,\tilde{Q}_{\theta}),\tilde{\epsilon}(t)  \sim \mathcal{N}(0,\tilde{Q}_x),$ and $\zeta(t)  \sim  \mathcal{N}(0,R_y),$ are zero mean white noise independent of $x(t),\theta(t)$ and $y(t)$.
The subscript $t$ indicates vectors that vary through time.

\textbf{The partial measurement problem:}
Ideally, states $x(t)$ would be directly measured, and thus appear as an element in $y(t)$. In practice, some of these states could influence $y(t)$ only indirectly by acting on other measurable states, where $d_y < d_x$. That is when classical training fails. 
In this work, we are interested in learning the unknown dynamics  $a(x(t),\theta(t))$ governing unmeasured states $x_h(t)$, where  
\begin{equation}
    \dot{x}_h(t)=a(x(t),\theta(t)) + \tilde{\epsilon}_h(t)
    \label{eq:hidden_dynamics}
\end{equation} 
where $x_h(t) \subset x(t)$ and $ \tilde{\epsilon}_h(t)  \subset \tilde{\epsilon}(t)$.This scenario poses further challenges over the estimation process since the recovery of latent states can be compromised. 
\paragraph{Identifiability of latent states:}
The task of recovering latent states $x(t)$ from a sequence of measurements and inputs $\mathcal{D}_N \overset{\Delta}{=} \{u(0),y(0),\dots, u(N-1),y(N-1)\}$ depends on the relationship between measurements and latent states. This problem gets even more complicated when model parameters need to be simultaneously estimated. 
Recent works define different forms of identifiability to analyze scenarios where latent states and parameters can be properly estimated~\cite{wieland2021structural}.
% Recent works discussed the identifiability of parameters and states providing definitions where [with words WHAT those definitions do] \citep{wang2021posterior,wieland2021structural}.
Specifically, they define identifiability of latent variables $x(t)$ as follows:

% rests on our ability to identify the latent state $x(t)$ form the $\mathcal{D}_N$.
\begin{definition}[State Identifiability]\label{def:dist_states}
We say that  latent variable $x(t_a)$  is identifiable given a parameter value $\hat{\theta}(t)$ and a measurement sequence  $y(t) \in\mathcal{Y}\subset\mathbb{R}^{d_y}$ if~\citep{wang2021posterior,wieland2021structural}
\begin{equation}
        x(t_a)\neq x(t_b) \implies h(x(t_a)) \neq h(x(t_b)) \,.
\end{equation}
\end{definition}
Although it is extremely difficult to provide formal guarantees, it makes sense that if for a given parameter $\hat{\theta
}(t)$, $h(x(t_a))=h(x(t_b))$, then obtaining an estimator for true state $x(t)$ becomes extremely challenging if not unfeasible. Since the proposed approach in this paper relies on estimating unmeasured states to learn model parameters, state identifiability is preferred.

To enforce latent variable identifiability, it is sufficient to ensure that the measurement function $h$ is an injective function \citep{wang2021posterior} for all $\theta$. Nevertheless, constructing an injective measurement function requires that $d_y \geq d_x$ \citep{wang2021posterior}, which is not feasible when dealing with the partial measurement problem where $d_y<d_x$.  
Moreover, for models with a high number of connected parameters, such as neural networks, enforcing identifiabilities can  be challenging \citep{wieland2021structural} and latent identifiability as defined in  Definition~\ref{def:dist_states} is not always guaranteed, especially when the number of measured states is less than the number of latent states. Note that a latent variable may be identifiable in a model given one dataset but not another, and at one $\theta$ but not another \citep{wang2021posterior}.
However, one could argue that one way to impose state identifiability is to re-parameterize the model~\citep{wieland2021structural} and incorporate prior knowledge regarding the relationship of states, focusing on achieving the properties stated in Definition~\ref{def:dist_states}.

\renewcommand\twocolumn[1][]{#1}%
\begin{figure*}
\centering
%\vspace{-0.6cm}
% \begin{multicols}{2}
%\vspace*{-0.001in}
   {\includegraphics[width=0.2\textwidth]{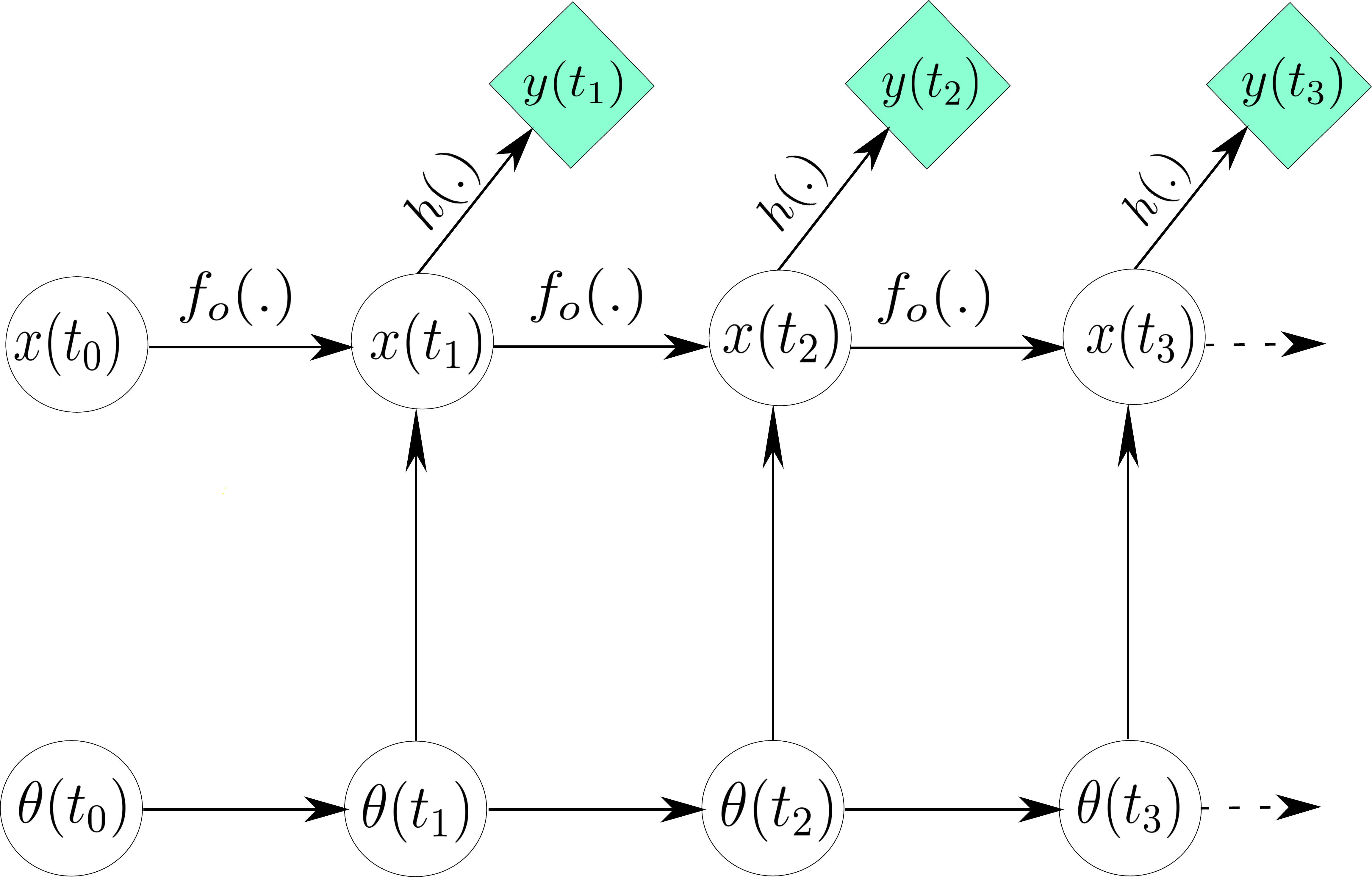}}
\qquad\quad
% \adjustbox{trim=0.32cm 0.3cm 4.5cm 0cm}{%\vspace{-0.35cm}
\includegraphics[width=0.4\textwidth]{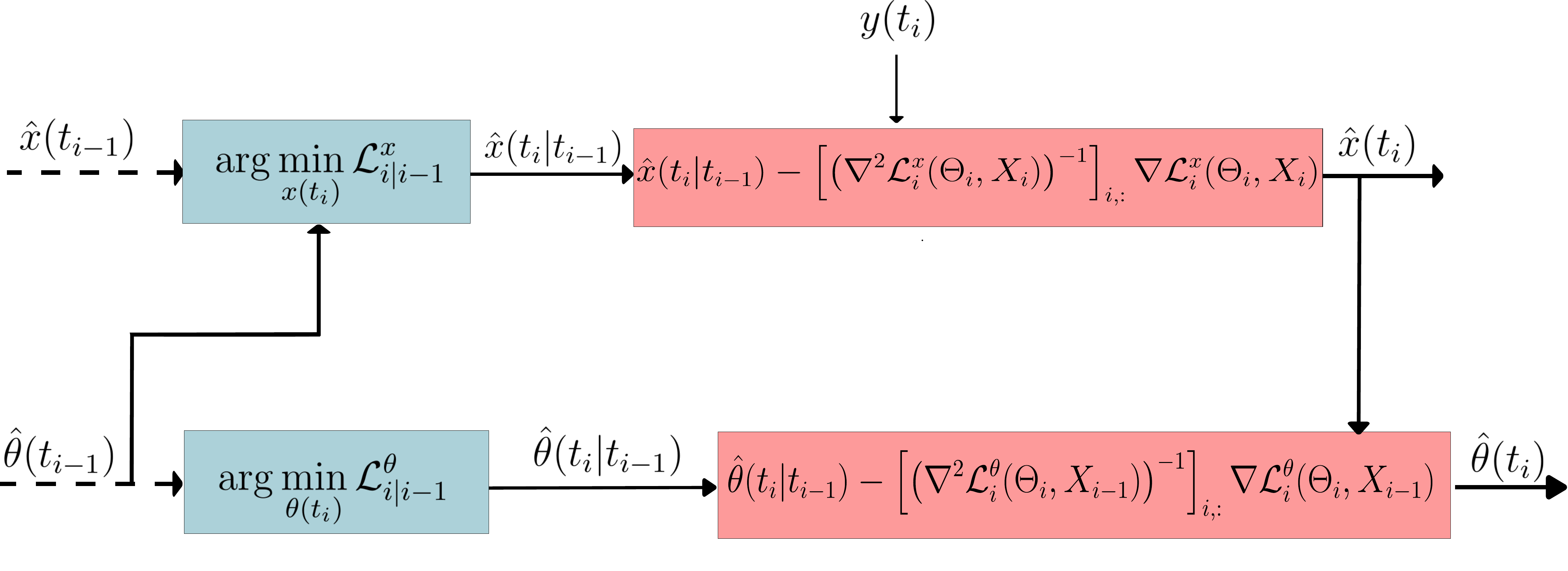}
% }
% \end{multicols}
%\vspace{-0.3cm}
\caption{The generative model (left panel), and one step of the proposed optimization strategy (right panel).}
\label{fig:overview_fig}
%%\vspace*{0.2in}
\end{figure*}

%\vspace{-0.3cm}
\subsection{Discrete Generative model}
%\vspace{-0.15cm}
In the continuous model presented in (\ref{eq:continuous_model}),a continuous-time description for the system states and parameters is assumed even though the measurements are recorded at discrete time points. 
Moreover, a function $a(x(t),\theta(t))$ was defined to describe the NODE governing the dynamics of the unmeasured states, and $u(t)$ described the external control inputs. In what follows, we will omit to use $a$ and $u(t)$ for notation simplicity, where $f(x(t),u(t),a(x(t),\theta(t))$ will be denoted by $f(x(t),\theta(t))$.
The discretization of states $x(t)$ and parameters $\theta(t)$  can therefore be expressed as time integration of (\ref{eq:continuous_model}) using Euler-Maruyama method \citep{sarkka2023bayesian} with uniform time step $\Delta_t=t_i-t_{i-1}$:
\begin{equation}
    \begin{aligned}
    x(t_i)&\!=\!x(t_{i-1}) + \int_{t_{i-1}}^{t_i}f(x(t),\theta(t))dt + \int_{t_{i-1}}^{t_i}\tilde{\epsilon}(t)dt\\
          x(t_i) &\!=\!x(t_{i-1})+ \Delta_t f(x(t_{i-1}),\theta(t_{i-1})) +\Delta_t\tilde{\epsilon}(t_{i-1})
    \end{aligned}
    \label{eq:trajectory_generation}
\end{equation}
Hence we define the following equation: 
\begin{equation}
    \!\!f_o(x(t_{i-1}),\theta(t_{i-1}))\!=\! x(t_{i-1}) + \Delta_t f(x(t_{i-1}),\theta(t_{i-1}))
    \label{eq:fo}
\end{equation} 
In a similar fashion of states $x(t)$, we discretize the parameters $\theta(t)$ and define the following equation:
\begin{equation}
    \begin{aligned}
    \theta(t_{i}))&=\theta(t_{i-1}) +   \int_{t_{i-1}}^{t_i}\tilde{\nu}(t)dt = \theta(t_{i-1}) + \nu(t)
    \label{eq:go}
    \end{aligned}
\end{equation}
Based on the continuous model presented in (\ref{eq:continuous_model}) and state discretization presented in (\ref{eq:trajectory_generation}, \ref{eq:fo},\ref{eq:go}), we present the discrete time evolution of the system states and parameters by the following discrete  generative model:
\begin{equation}
    \begin{aligned}
        \theta(t_i)&=\theta(t_{i-1}) + \nu(t)\\
        x(t_i)&= f_o(x(t_{i-1}),\theta(t_{i-1}))+ \epsilon(t)\\
        y(t_i)&=h(x(t_i)) + \zeta(t)\,.
        \label{eq:generative_model}
    \end{aligned}
\end{equation}
where $\nu(t)\!\sim\!\mathcal{N}(0,Q_{\theta}), \epsilon(t) \!\sim\!\mathcal{N}(0,Q_x),$ and $\zeta(t)\!\sim\!\mathcal{N}(0,R_y)$, with $Q_{\theta}=\Delta_t\tilde{Q}_{\theta}$ and $Q_x=\Delta_t\tilde{Q}_x$.


%\vspace{-0.3cm}
\section{Method} \label{sec:method}
%\vspace{-0.2cm}

The proposed approach finds the model parameters $\theta(t)$ of hidden Neural ordinary differential equation $a(x,\theta)$ describing $x_h(t) \in x(t) $ and latent states $x(t)$ of dynamical system given a dataset $\mathcal{D} \triangleq \{u(t_0),y(t_0),\dots, u(t_{N-1}),y(t_{N-1})\}$ of discrete measurements and control inputs when $x(t)$ is partially measured, that is $x_h(t) $ is unmeasured. We formulate the problem of estimating $x(t)$ and $\theta(t)$ as an optimization problem that is similar to \citep{humpherys2012fresh}, where we optimize a cost function $\mathcal{L}$ given a probabilistic discrete model (\ref{eq:generative_model}), exploiting the link between the second-order Newton's method and the Kalman filter. The cost function $\mathcal{L}$ is updated and solved sequentially to find latent states $x(t)$ and model parameters $\theta(t)$ in a unified framework. Our approach assumes state identifiability which implies that latent states $x(t)$ are recoverable from measurements $y(t)$. In this context, we break the optimization steps into two concerning optimization with respect to $x(t)$ and $\theta(t)$.
% \textcolor{black}{This procedure associated with the distinguishability of the states lead to new update rules capable of optimizing both states and parameters. If the states are not distinguishable, finding parameters $\theta(t)$ becomes unfeasible.maybe delete? }

%A solution ${x^*,\theta^*}$ of \ref{eq:full_optimization_function} can be computed by different unconstrained nonlinear programming solvers and BPTT. 

% \subsection{Model}
%\vspace{-0.03cm}
\subsection{Sequential Newton Derivation}
%\vspace{-0.075cm}
We denote by $\Theta_N=[\theta(t_0),\dots,\theta(t_N)]$ and $X_N=[x(t_0),\dots,x(t_N)]$ to 
be the set of latent states sampled at $t_0,t_1,\dots,t_N$.
To train the model, we optimize $(\Theta_N,X_N)$ to minimize a quadratic cost function starting from initial $\{x(t_0),\theta(t_0)\}$ using a collection of combined measurement and input sequences $\mathcal{\bm{D}} $. A physics informed loss term is employed in our cost function to help identify hidden states $x_h$,   where the cost function is defined as:
%\vspace{-0.05cm}
\begin{align} \label{eq:full_optimization_function}
&\mathcal{L}_N(\Theta_N,X_N)= \frac{1}{2}\sum_{i=1}^{N}  \overbrace{\lVert x(t_i) - f_o(x(t_{i-1}),\theta(t_{i-1}))\rVert^{2}_{Q_x^{-1}}}^{\text{\color{red} Physics Informed loss}} \nonumber \\
&+ \underbrace{\lVert y(t_i) - h(x(t_i)) \rVert^{2}_{R_y^{-1}}}_{\text{\color{blue} Data driven loss}} + \underbrace{\lVert \theta(t_i) -\theta(t_{i-1})\rVert^{2}_{Q_{\theta}^{-1}}}_{\text{\color{orange}Regularization term}} \,. 
\end{align}
%\vspace{-0.08cm}
where $Q_x$, $R_y$ and $Q_\theta$ are known positive definite matrices corresponding to latent states, measurement and parameter uncertainty respectively , and $\|a - b\|^2_{A^{-1}} = (a-b)^T A^{-1}(a-b)$.
% $. that are equivalent to A in the following weighted norm equation (\ref{eq:weighted})  
% \begin{equation}
%      \|a - b\|^2_{A^{-1}} = (a-b)^\top A^{-1}(a-b).
%      \label{eq:weighted}
% \end{equation}
As the Hessian's inverse is in general intractable, finding optimal solution $(\Theta_N^*,X_N^*)$ using the second order Newton method over the whole data set of size $N$ is unfeasible. For this reason, we resort to a sequential strategy by introducing a modified quadratic function $\mathcal{L}_{i}(\Theta_i,X_i)$. 
%For this reason, we adapt the optimization function (\ref{eq:full_optimization_function}) in order to be suited for sequential optimization: 
Let us re-write the cost function at time $t_i$ as:
% {\small
% \begin{equation} 
%\vspace{-0.25cm}
\begin{align}
\label{eq:divided_optimization_function}
&\mathcal{L}_{i}(\Theta_i,X_i)  =\mathcal{L}_{i-1}(\Theta_{i-1},X_{i-1})\nonumber \\ &   + \frac{1}{2}\lVert x(t_i) - f_o(x(t_{i-1}),\theta(t_{i-1})) \rVert^{2}_{Q_x^{-1}}   \\
& + \frac{1}{2}\lVert y(t_i) - h(x(t_i)) \rVert^{2}_{R_y^{-1}}  + \frac{1}{2}\lVert \theta(t_i) -\theta(t_{i-1}) \rVert^{2}_{Q_{\theta}^{-1}} \nonumber
\end{align}
%\vspace{-0.45cm}
% \end{equation}
% }

%
where $\mathcal{L}_{i-1}(\Theta_{i-1},X_{i-1})$ and $\mathcal{L}_{i}(\Theta_i,X_i)$ are the cost functions at times $t_{i-1}$ and $t_i$, respectively; $\Theta_i=[\theta(t_0),\dots,\theta(t_i)]$ and $X_i=[x(t_0),\dots,x(t_i)]$. In the sequential optimization paradigm, $\Theta_{i-1}$ and $X_{i-1}$ are assumed known and at the $i$-th optimization step is performed only with respect to $\{\theta(t_i),x(t_i)\}$.
When $\{\theta(t_i),x(t_i)\}$ are determined jointly such as in \citep{humpherys2012fresh}, the optimization process will suffer from vanishing gradients under partial measurements, see Appendix~\ref{app:vanishingGrad}. However, if $x(t_i)$ is identifiable, we can circumvent the vanishing gradient problem by first optimizing with respect to $x(t_i)$ and then $\theta(t_i)$. To improve identifiability of latent states $x(t)$, we employ a physics informed term in the cost function described in \ref{eq:full_optimization_function} and combine it with the alternating optimization approach proposed below, that optimizes $x(t)$ and $\theta(t)$ separately. This will allow us to circumvent the partial observability problem and enable the use of an estimate of the unmeasured state in training.
To do so, we break the optimization function  (\ref{eq:divided_optimization_function}) into four alternating optimization procedures aiming at finding $\hat{x}(t_i)$ and then finding $\hat{\theta}(t_i)$ that minimizes (\ref{eq:divided_optimization_function}) given $\hat{x}(t_i)$. 

Let us begin by defining two intermediate optimization functions $\mathcal{L}_{i|i-1}^{x}$ and $\mathcal{L}_{i|i-1}^{\theta}$ in (\ref{eq:divided_optimization_function_x}) and (\ref{eq:divided_optimization_function_theta}) respectively as follows:
%\vspace{-0.25cm}
{\small
\begin{align}
\label{eq:divided_optimization_function_x}
&\mathcal{L}_{i|i-1}^{x}(\Theta_i,X_i) = \ \mathcal{L}_{i-1}(\Theta_{i-1},X_{i-1})\\
& \!\!\! + \frac{1}{2}\lVert x(t_i) \! - \! f_o(x(t_{i-1}),\theta(t_{i-1})) \rVert^{2}_{Q_x^{-1}} + \frac{1}{2}\lVert \theta(t_i) -\theta(t_{i-1}) \rVert^{2}_{Q_{\theta}^{-1}} \nonumber
\end{align}
% \begin{align}
% \label{eq:divided_optimization_function_x}
% &\mathcal{L}_{i|i-1}^{x}(\Theta_i,X_i) = \ \mathcal{L}_{i-1}(\Theta_{i-1},X_{i-1})\nonumber\\& \!\!\! + \frac{1}{2}\lVert x(t_i) \! - \! f_o(x(t_{i-1}),\theta(t_{i-1})) \rVert^{2}_{Q_x^{-1}} + \frac{1}{2}\lVert \theta(t_i) -\theta(t_{i-1}) \rVert^{2}_{Q_{\theta}^{-1}} 
% \end{align}
}
%\vspace{-0.1cm}
% and 
\begin{equation} 
{\small
\begin{aligned}
\mathcal{L}_{i|i-1}^{\theta}(\Theta_i,X_{i-1})&= \! \mathcal{L}_{i-1}(\Theta_{i-1},X_{i-1})+\frac{1}{2} \lVert \theta(t_i) - \theta(t_{i-1}) \rVert^{2}_{Q_{\theta}^{-1}}  \,.
\end{aligned}
}
\label{eq:divided_optimization_function_theta}
\end{equation} 
We proceed by optimizing  (\ref{eq:divided_optimization_function_x}) for $x(t_i)$ and (\ref{eq:divided_optimization_function_theta}) for $\theta(t_i)$, yielding the respective solutions below:
%\textcolor{black}{by setting the gradients of (\ref{eq:divided_optimization_function_x}) and (\ref{eq:divided_optimization_function_theta}) to zero. To do so we fix $x$ at $\hat{x}(t_i|t_{i-1})$ and $\theta$ at $\hat{\theta}(t_i|t_{i-1})$: maybe delete}
\begin{equation}
\begin{aligned}
   \hat{\theta}(t_i|t_{i-1})&=\hat{\theta}(t_{i-1})\\
   \hat{x}(t_i|t_{i-1})&= f_o(\hat{x}(t_{i-1}),\hat{\theta}(t_{i-1}))\,.
   \label{eq:predict}
\end{aligned}
\end{equation}
Next, we define the two optimization functions responsible for the update steps for states and parameters. Specifically, we define $\mathcal{L}_i^{x}$ as: 
% $\mathcal{L}_i^{\theta}$ in (\ref{eq:L_x})  and (\ref{eq:L_theta}) to be optimized with respect to $x(t_i)$ and $\theta(t_i)$, respectively. Thus, $\mathcal{L}_i^{x}$
\begin{equation}
\mathcal{L}_i^{x}(\Theta_i,X_i)=\mathcal{L}_{i|i-1}^{x}(\Theta_i,X_i)+ \lVert y(t_i) - h(x(t_i)) \rVert^{2}_{R_y^{-1}}
\label{eq:L_x}
\end{equation}
to be optimized with respect to ${x}(t_i)$ by minimizing $\mathcal{L}_i^{x}$ given intermediate values of equation (\ref{eq:predict}) where: 
\begin{equation}
    % \hat{x}(t_i)=\hat{x}(t_i|t_{i-1}) -row_i\left(\nabla^2 \mathcal{L}_{i}^{x}(\Theta_i,X_i)\right)^{-1}\nabla \mathcal{L}_{i}^{x}(\Theta_i,X_i) 
    \hat{x}(t_i)=\hat{x}(t_i|t_{i-1}) -\left[\left(\nabla^2 \mathcal{L}_{i}^{x}(\Theta_i,X_i)\right)^{-1}\right]_{i,:}\nabla \mathcal{L}_{i}^{x}(\Theta_i,X_i) \,.\nonumber
    % \label{eq:update_x_newton} 
\end{equation}
The solution to the above problem is given by given by (\ref{eq:x_update}). 
% The resulting optimal variable $\hat{x}(t_i)$ is given by (\ref{eq:x_update}). 
Equivalently, we define the update optimization function $\mathcal{L}_i^{\theta}$ as:
{\small
\begin{align}
&\mathcal{L}_i^{\theta}(\Theta_i,X_i)=\mathcal{L}_{i|i-1}^{\theta}(\Theta_i,X_{i-1})\\
&+ \lVert x(t_i) - f_o(x(t_{i-1}),\theta(t_{i-1}))  \rVert^{2}_{Q_x^{-1}} + \lVert y(t_i) - h(x(t_i)) \rVert^{2}_{R_y^{-1}} \nonumber
\label{eq:L_theta}    
\end{align}
}
% \begin{equation}
% {\small
% \begin{aligned}
% &\mathcal{L}_i^{\theta}(\Theta_i,X_i)=\mathcal{L}_{i|i-1}^{\theta}(\Theta_i,X_{i-1})\\
% &+ \lVert x(t_i) - f_o(x(t_{i-1}),\theta(t_{i-1}))  \rVert^{2}_{Q_x^{-1}} + \lVert y(t_i) - h(x(t_i)) \rVert^{2}_{R_y^{-1}} 
% \label{eq:L_theta}
% \end{aligned}
% }
% \end{equation}
\!to be optimized with respect to ${\theta}(t_i)$ by minimizing $\mathcal{L}_i^{\theta}$ given intermediate values of equation (\ref{eq:predict}) and (\ref{eq:x_update}) as follows:
{\small
\begin{equation*}
    \hat{\theta}(t_i)=\hat{\theta}(t_i|t_{i-1}) -\left[\left(\nabla^2 \mathcal{L}_{i}^{\theta}(\Theta_i,X_{i-1})\right)^{-1}\right]_{i,:}\nabla \mathcal{L}_{i}^{\theta}(\Theta_i,X_{i-1})
    \label{eq:update_theta_newton}
\end{equation*}
}
The resulting optimal variable $\hat{\theta}(t_i)$ is given by (\ref{eq:theta_update}). The procedure is repeated until $t_i=t_N$. We present our main result in the following theorem:
\begin{theorem}\label{theorem1}
   % Starting from known
   Given 
   $\hat{\theta}(t_{i-1}) \in \hat{\Theta}_{i-1}$ and $\hat{x}(t_{i-1})  \in \hat{X}_{i-1}$, and known $P_{\theta_{i-1}}\in R^{d_{\theta}\times d_{\theta}}$ and $P_{x_{i-1}} \in R^{d_x \times d_x}$, 
   the recursive equations for computing $\hat{x}(t_i)$ and $\hat{\theta}(t_i)$ that minimize (\ref{eq:divided_optimization_function}) are given by the following:
    {\small
    \begin{align} \label{eq:x_update} 
       \hat{x}(t_{i})&\!=\!f_o(\hat{x}(t_{i-1}),\hat{\theta}(t_{i-1}))   \\ & \!\hspace{-0.5cm}-\!P_{x_i}^{-}H_i^T\!\!\left(H_iP_{x_i}^{-}H_i^T \!+\! R_y\right)^{\!-1}\!\!\left[h\!\left(f_o(\hat{x}(t_{i-1}),\hat{\theta}(t_{i-1}))\right)\!-\!y(t_i)\right] \nonumber 
         \\
    % \end{equation}
    % \begin{equation}
       \hat{\theta}(t_i)&\!=\!\hat{\theta}(t_{i-1})\label{eq:theta_update} \!-\! P_{\theta_i}^{-}F_{\theta_{i-1}}^T                       \left[f_o(\hat{x}(t_{i-1}),\hat{\theta}(t_{i-1}))\!-\!\hat{x}(t_{i})\right]  
    \end{align}
    }
   
\noindent with $P_{\theta_i}^{-}$, $P_{x_i}^{-}$ being intermediate matrices and $P_{\theta_i}^{}$ and $P_{x_i}^{}$  being the lower right blocks of  $(\nabla^2 \mathcal{L}_{i}^{\theta})^{-1}$ and $(\nabla^2 \mathcal{L}_{i}^{x})^{-1}$ respectively:
% {\small
% \begin{equation}
%     \begin{aligned}
%     P_{\theta_i}^{-}&=P_{\theta_{i-1}} -P_{\theta_{i-1}}F_{\theta_{i-1}}^{T}\left(Q_x+F_{\theta_{i-1}}P_{\theta_{i-1}}F_{\theta_{i-1}}^{T}\right)F_{\theta_{i-1}}P_{\theta_{i-1}}\\
%     P_{x_i}^{-}&=F_{x_{i-1}}P_{x_{i-1}}F_{x_{i-1}} +Q_x \\ 
%        P_{x_i}&=P_{x_i}^{-}[I+ H_i\left(R_y-H_iP_{x_i}^{-}H_i^{T}\right)H_{i}P_{x_i}^{-}]\\
%        P_{\theta_i}&=Q_{\theta}+P_{\theta_i}^{-}
%     \end{aligned}     
% \end{equation}
% }
\begin{align}
    P_{\theta_i}^{-}&=P_{\theta_{i-1}} -P_{\theta_{i-1}}F_{\theta_{i-1}}^{T}\left(Q_x+F_{\theta_{i-1}}P_{\theta_{i-1}}F_{\theta_{i-1}}^{T}\right)F_{\theta_{i-1}}P_{\theta_{i-1}}\nonumber\\
    P_{x_i}^{-}&=F_{x_{i-1}}P_{x_{i-1}}F_{x_{i-1}} +Q_x \\ 
       P_{x_i}&=P_{x_i}^{-}[I+ H_i\left(R_y-H_iP_{x_i}^{-}H_i^{T}\right)H_{i}P_{x_i}^{-}] \nonumber\\
       P_{\theta_i}&=Q_{\theta}+P_{\theta_i}^{-} \nonumber
\end{align}
with $H_i,F_{x_{i-1}}$, and $F_{\theta_{i-1}}$ being the jacobians of the vector fields $h$ and $f_o$  at $\hat{x}(t_i|t_{i-1}),\hat{x}(t_{i-1})$ and $\hat{\theta}(t_{i-1})$: 

$H_i=\frac{\partial h(\hat{x}(t_i|t_{i-1}))}{\partial \hat{x}(t_i|t_{i-1})}$, $F_{x_{i-1}}=\frac{\partial f_o(\hat{x}(t_{i-1}),\hat{\theta}(t_{i-1}))}{\partial \hat{x}(t_{i-1})}$  \quad  and $F_{\theta_{i-1}}=\frac{\partial f_o(\hat{x}(t_{i-1}),\hat{\theta}(t_{i-1}))}{\partial \hat{\theta}(t_{i-1})}$ 

\end{theorem}

The proof of Theorem \ref{theorem1} is provided in Appendix~\ref{app:proof_theo1}.
As a consequence of Theorem (\ref{theorem1}), $\hat{x}(t_i)$ is computed according to (\ref{eq:x_update}) using $\hat{\theta}(t_{i-1})$. $\hat{\theta}(t_i)$ is computed afterwards according to (\ref{eq:theta_update}) using $\hat{x}(t_i)$ that was previously found in (\ref{eq:x_update}). This alternating procedure between $x(t_i)$ and $\theta(t_i)$ is explained in the right panel of Figure \ref{fig:overview_fig}, which depicts the four alternate optimization steps performed for each iteration $t_i$. The computational complexity of the proposed approach is detailed in~Appendix \ref{app: Complexity}. An epoch has a complexity of $\mathcal{O}(N(d_x^3 + 2d_{\theta}^2d_x + 2d_{\theta}d_x^2))$. Under the assumption that $d_\theta \gg d_x$ the complexity becomes $\mathcal{O}(N(2d_{\theta}^2d_x + 2d_{\theta}d_x^2))$. During testing, however, the complexity becomes $\mathcal{O}(d_\theta)$ per step if integrating the learned mean vector field.
%\textcolor{red}{\subsection{Obtaining initial condition from historical data}\label{historical data}
%Obtaining initial conditions $x(t_0)$ during test time is often challenging.  

%However, the proposed recursive framework can easily provide an estimate of the initial condition if historical data $\mathcal{D}_{\cal H} \triangleq \{u(t_{-N}),y(t_{-N}),\ldots, u(t_0),y(t_0)\}$ is available as described in~\eqref{eq:optim_x0} in Appendix \ref{x0_appendix}. Thus, given the optimal model after training $\theta^*$ we can exploit the update equation for the states, see (\ref{eq:theta_update}), to provide $\hat{x}(t_0)$.}

%\vspace{-0.2cm}
\section{Experiments}
%\vspace{-0.2cm}
The performance of the proposed approach is assessed in comparison to state-of-the-art model learning methods on several challenging nonlinear simulations and real-world datasets. 
We employed five different dynamical models to demonstrate the effectiveness of the proposed approach. For each dynamical model, we assumed that we don't have parts of the governing dynamics available, and replaced them with a neural network.  Euler integrator is used as the ODE solver for efficiency and fast computation speed.

% \textcolor{red}{Since the proposed mechanism rests on determining unmeasured latent states from measured ones, successful learning of the model relies on the identifiability of latent states as defined in Definition (\ref{def:dist_states}). To ensure that, we assume partial knowledge of system ODE's}\textcolor{black}{, where the dynamics of measured states are known.}

As benchmark methods, we considered five other well-established techniques for dynamical machine learning, namely NODE~\citep{chen2018neural}, NODE-LSTM~\citep{chen2018neural}, SPINODE~\citep{o2022stochastic}, RM \citep{buisson2022recognition} and PR-SSM \citep{doerr2018probabilistic}. 
We denote that NODE-LSTM is a NODE with an LSTM network, and SPINODE is a Stochastic Physics Informed NODE.
Currently, no code is available for the model learning frameworks presented in \citep{eleftheriadis2017identification}. Moreover, the available code related to the works in  \citep{mchutchon2015nonlinear,ialongo2019overcoming} could be modified to account for the partial measurement scenario. However, these algorithms become computationally unfeasible for medium and large datasets \citep{doerr2018probabilistic}. For that reason, we were not able to benchmark against these approaches. 
We emphasize that modifying the above-mentioned methods to either account for the ODE structure or make them computationally tractable is out of the scope of this paper. This also applies to the PRSSM method. Nevertheless, for the sake of providing comparative results, we still include results using PR-SSM which is computationally more efficient than other Gaussian process-based models but does not account for the ODE structure.  
% Reproducing these methods is not straightforward and outside the scope of this work. 

 % $$\mathrm{nRMSE}=\frac{\sqrt{\frac{1}{n}\sum_{i=1}^{n}(x(t_i) - \hat{x}(t_i))^2}}{\max(x(t)) - \min(x(t))},$$
The benchmark results are summarized in Table \ref{tab:RMSEs} which represents normalized Root Mean Square Error (nRMSE) values for each model and method. In Figs. \ref{fig:hh_result}-\ref{fig:EMPS_result} we compare the benchmark methods, and our proposed method. All results were obtained with learned mean vector field integrated over time. Each subfigure represents the dynamics of a single state and contains ODE solutions for each method. We computed nRMSE using $\mathrm{nRMSE}=\sqrt{\frac{1}{n}\sum_{i=1}^{n}(x(t_i) - \hat{x}(t_i))^2} / \bigr[ \max(x(t)) - \min(x(t)) \bigr],$ where $\hat{x}(t_i)$ and $x(t_i)$ are the estimated and true states at time $t_i$, respectively, and $n$ is the number of data points.
Moreover, the learning curves for the proposed approach on each example are presented in Appendix~\ref{app: Learning Curves}. 

%\vspace{0.3cm}
\begin{table*}[ht]
\centering
% \small
\renewcommand{\arraystretch}{0.7}
\setlength{\tabcolsep}{3.3pt}
%\vspace{0.2cm}
%%\vspace{-0.2cm}
\small
\begin{tabular}{p{6.5cm}|p{1.6cm}p{1.8cm}p{1.6cm}p{2cm}p{1.6cm}}
\toprule
Methods &  HH model  & Yeast  Glyco. & Cart-pole & Harmonic Osc. & EMPS\\ \bottomrule
\toprule
RM \citep{buisson2022recognition}     & $2.39 \cdot 10^{-1}$ & $ 6.30 \cdot 10^{-1}$ & $ 1.06\cdot 10^{0}$ & $ 2.36 \cdot 10^{-2}$ & $ 6.20 \cdot 10^{-1}$\  \\ \midrule
PR-SSM \citep{doerr2018probabilistic} & $4.05 \cdot 10^{-1}$ & $ 1.59\cdot 10^{0}$  & $ 1.52\cdot 10^{0}$ & $ 1.21 \cdot 10^{0}$  & $ 4.05 \cdot 10^{1}$ \  \\ \midrule
SPINODE \citep{o2022stochastic} & $ 7.68 \cdot 10^{-1}$ & $ 4.98 \cdot 10^{-2}$  & $3.01 \cdot 10^{0}$ & $ 4.34 \cdot 10^{-1}$ & $4.30 \cdot 10^{5}$ \  \\ \midrule
NODE \citep{chen2018neural}  & $ 7.03 \cdot 10^{1}$ & $ 3.74 \cdot 10^{-1}$ & $ 2.84 \cdot 10^{-1}$ & $ 4.65 \cdot 10^{-1}$ &  $ 1.65 \cdot 10^{0}$ \  \\ \midrule
NODE-LSTM \citep{chen2018neural} & $ 3.87 \cdot 10^{1}$ & $ 3.09 \cdot 10^{-1}$ & $2.90 \cdot 10^{-1}$ & $ 4.60 \cdot 10^{-1}$ & $3.45 \cdot 10^{0}$ \  \\ \midrule
Proposed Approach                      & $\mathbf{1.54 \!\cdot\! 10^{-1}}$ & $\mathbf{3.39   \!\cdot\! 10^{-2}}$ & $ \mathbf{9.41\!\cdot\! 10^{-3}}$ & $ \mathbf{5.08 \!\cdot\! 10^{-3}}$ & $ \mathbf{9.50 \!\cdot\! 10^{-2}}$\\ \toprule
\end{tabular}
%\vspace{-0.3cm}
\caption{Comparison of nRMSE values for different dynamical models and methods.}
\label{tab:RMSEs}
\end{table*}

% \begin{figure}[htb]
% \centerline{\includegraphics[trim={0 0 0 0.1cm},clip,width=\columnwidth]{figure1.png}} 
% \caption{}
% \label{fig:all_results}
% \end{figure}

% \textcolor{black}{[compare with different initial conditions]}

%\vspace{-0.3cm}
\subsection{Hodgkin-Huxley Neuron Model}
% \vspace{-0.2cm}

%
%
\begin{figure}
% \centerline{\includegraphics[trim={0 0 0 0.1cm},clip,width=\myfigurewidth cm]{figure_EMPS.pdf}}
\centering
% \vspace{-1.1cm}
\includegraphics[width=0.38\textwidth]{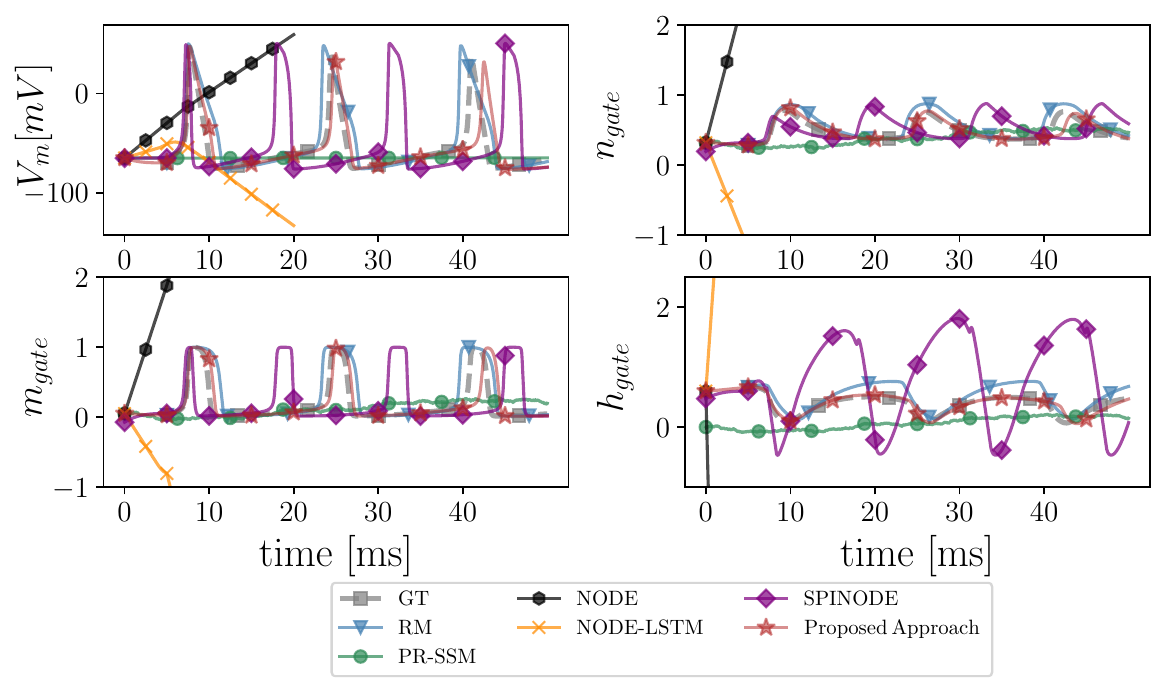}
% \vspace{-0.3cm}
% \captionsetup{font=small}
\caption{Learned state trajectories of HH model after training with RM, PR-SSM, NODE, NODE-LSTM, SPINODE methods and our proposed approach. Results are compared to ground truth ODE system trajectory labeled as GT. The proposed approach  is capable of discerning the true trajectory for the unmeasured state $h_{gate}$.}
\label{fig:hh_result}
\end{figure}

The Hodgkin-Huxley (HH) Neuron Model \citep{Hodgkin_1952} is an ODE system that describes the membrane dynamics of action potentials in neurons, which are electrical signals used by neurons to communicate with each other. The model has four states: $\dot{V}_m$ is the membrane potential, $n_{gate}$, $m_{gate}$, and $h_{gate}$ are gating variables controlling the membrane's ionic permeability. The equations governing the ODE system are provided in \eqref{v_m_states}-\eqref{eq:dq} of the Appendix \ref{HH_appendix}. We train our recursive model with the assumption that Eq. (\ref{eq:dq}) governing dynamics of $h_{gate}$ is unknown and its corresponding state is not measured, i.e., $y(t_i)= (V_m(t_i), n_{gate}(t_i), m_{gate}(t_i))$. We replace the dynamics describing $\dot{h}_{gate}(t)$ by a neural network consisting of  three feed-forward layers for all the benchmark methods except NODE-LSTM where we used three LSTM layers . The first layer is a 20 units layer followed by an Exponential Linear Unit ($ELU$) activation function, the second layer is also a 20 unit layer followed by a $tanh$ activation function. The last layer consists of 10 units with a $sigmoid$ activation function.
We generate the dataset by applying a constant control input $u(t_i)$ to the HH model described in (\ref{v_m_states})-(\ref{eq:dq}) for 50000 time steps with $dt=10^{-3}s$ and by collecting measurements and inputs  $\mathcal{D} \triangleq \{u(t_0),y(t_0),\dots, u(t_{N-1}),y(t_{N-1})\}$.  We train our model on $\mathcal{D}$ with $P_{x_0}=10^{-2}I_{d_x}, P_{\theta_0}=10^{2}I_{d_{\theta}}$ $R_y =10^{-10}I_{d_y}, Q_x=10^{-5}I_{d_x}$ and $ Q_{\theta}=10^{-2}I_{d_{\theta}}$. At the beginning of each epoch, we solve the problem (\ref{eq:optim_x0}) of the Appendix \ref{app:x0_appendix} to get the initial condition. Final optimal parameters $\hat{\theta}(t_N)$ and initial condition $\hat{x}(t_0)$ are saved and collected at the end of training.  
Fig. \ref{fig:hh_result} depicts the dynamics of the system $\hat{\theta}(t_N)$ generated according to the generative
model described in Eq (\ref{eq:trajectory_generation}) starting from initial condition $\hat{x}(t_0)$. The lower right panel demonstrates the superiority of the proposed model at learning  $h_{gate}$.

\begin{figure}
% \centerline{\includegraphics[trim={0 0 0 0.1cm},clip,width=\myfigurewidth cm]{figure_EMPS.pdf}}
\centering
%\vspace{-0.6cm}
\includegraphics[width=0.4\textwidth]{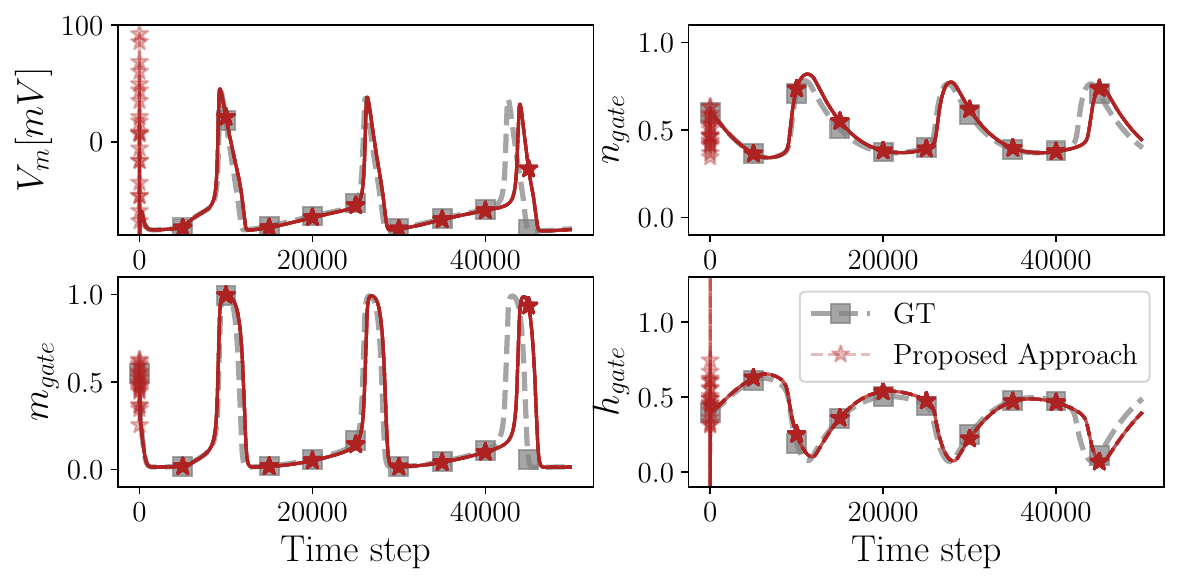}
%\vspace{-0.4cm}
% \captionsetup{font=small}
\caption{The proposed approach's results for unknown initial conditions. Initial conditions $\hat{x}(t_{100})$ were learned using the first 100 samples.}
\label{fig:figure_init}
\end{figure}
To demonstrate the robustness of the proposed approach to different dynamical regimes and showcase its capability of estimating accurate initial conditions, we perform an additional experiment. For this, we generate data $\mathcal{D}_T$ with $N=50,000$ samples using the HH model with different initial conditions from the ones used during training. From this data, we reserve the first 100 samples for learning the initial condition before performing integration for the remaining $49,900$ samples.   
Then, using the learned model $\hat{\theta}(t_N)$ and the procedure described in Appendix~\ref{app:x0_appendix} we obtained the initial condition $\hat{x}(t_{100})$ and obtained the proposed approach's solution. Figure~\ref{fig:figure_init} shows the evolution of the proposed approach attesting to its capability of both estimating accurate initial conditions and generalization to other dynamical regimes.

%change this to resize
\def\myfigurewidth{12}

% \vspace{-0.1cm}
% \vspace{0.3cm}
%\vspace{-0.3cm}
\subsection{Cart-pole System}
%\vspace{-0.25cm}
% We demonstrate the efficacy of the proposed approach in learning the nonlinear dynamics of the cart-pole system. 
The cart-pole system is composed of a cart running
on a track, with a freely swinging pendulum attached to it. The state of the system consists of the
cart’s position and velocity, and the pendulum’s angle and angular velocity, while a control input $u$ can be applied to the cart. We used the LQR \citep{prasad2011optimal}  algorithm to learn a feedback controller that swings the pendulum and balances it in
the inverted position in the middle of the track.
The equations governing the ODE system are provided in \eqref{eq:cartpole_state1}-\eqref{eq:cartpole_state4}  of the Appendix \ref{CP_appendix}.

\begin{figure}
% \centerline{\includegraphics[trim={0 0 0 0.1cm},clip,width=\myfigurewidth cm]{figure_EMPS.pdf}}
\centering
%\vspace{-0.9cm}
%\vspace{-0.2cm}
\includegraphics[width=0.4\textwidth]{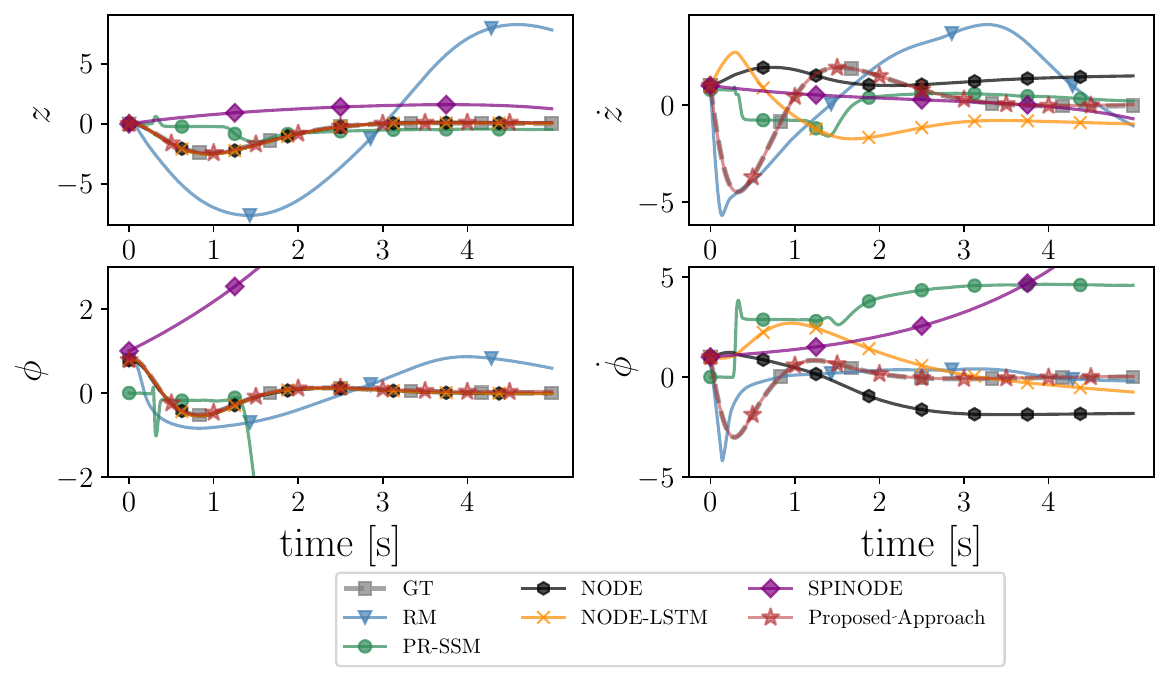}
%\vspace{-0.3cm}
%\captionsetup{font=small}
\caption{Learned state trajectories of the cart-pole system after training RM, PR-SSM, SPINODE, NODE, NODE-LSTM methods and the proposed approach. Results are compared to ground truth ODE system trajectory labeled as GT. We showed that the proposed approach can discern the true trajectory for the unmeasured states $\dot{z}$ and $\dot{\phi}$.}
\label{fig:cartpole_result}
\end{figure}

We train our recursive model with the assumption that we don’t know the equation corresponding to $\dot{\phi}$ governing dynamics of the cart-pole's angular rate. Therefore, we replace \eqref{eq:cartpole_state2} and \eqref{eq:cartpole_state4} with a two-layer LSTM neural network when training with NODE-LSTM and a two-layer feedforward neural network for the rest of the benchmarks, with $tanh$ activation function on each layer. We don’t measure cart-pole's velocity $\dot{z}(t_i)$ and angular rate $\dot{\phi}(t_i)$, i.e., $y(t_i) = [z(t_i), \phi(t_i)].$
We generate our dataset by applying LQR balancing controller to the cart-pole described in Eqs (\ref{eq:cartpole_state1})-(\ref{eq:cartpole_state4}) for 5000 time steps with $dt=10^{-3}s$ and by collecting measurements and inputs  $\mathcal{D} \triangleq \{u(t_0),y(t_0),\dots, u(t_{N-1}),y(t_{N-1})\}$. We train our model on $\mathcal{D}$ with $P_{x_0}=10^{-2}I_{d_x}, P_{\theta_0}=10^{2}I_{d_{\theta}}$ $R_y =10^{-10}I_{d_y}, Q_x=10^{-5}I_{d_x}$ and $ Q_{\theta}=10^{-2}I_{d_{\theta}}$. At the beginning of each epoch, we solve problem (\ref{eq:optim_x0}), Appendix \ref{app:x0_appendix}, to obtain initial conditions. The final optimal parameters $\hat{\theta}(t_N)$ and initial condition $\hat{x}(t_0)$ are saved and collected at the end of training.  
We qualitatively assess the performance of our model using the control sequence stored in $\mathcal{D}$ and optimal parameters $\hat{\theta}(t_N)$ according to the generative model described in~(\ref{eq:trajectory_generation}) starting from initial condition $\hat{x}(t_0)$.

% In Figure \ref{fig:cartpole_result}, we demonstrate the ability of the proposed approach to learn the underlying dynamics of the partially measured system compared to benchmark methods.  
%
%
Table \ref{tab:RMSEs} shows that the proposed approach outperforms the competing algorithms with nRMSE value that is  two to three orders of magnitude smaller when compared with competing methods. Analyzing the evolution of the latent states depicted in Figure~\ref{fig:cartpole_result}, we notice that our proposed approach provides state trajectories that match the ground truth (GT) while the other methods fail to capture the true trajectory. PR-SSM presents acceptable trajectories of $\dot{z}$ and $\dot{z}$ but fails to learn $\phi$ and $\dot{\phi}$ trajectories. On the other hand, RM presents acceptable trajectories of $\phi$ and $\dot{\phi}$ but fails to learn $z$ and $\dot{z}$ trajectories. Moreover, the NODE and NODE-LSTM successfully learn the measured $\phi$ and $z$ trajectories but fail to learn correct trajectories of the unmeasured states $\dot{\phi}$ and $\dot{z}$. 
SPINODE, RM, and PR-SSM estimated state trajectories are much more inaccurate than the one provided by our proposed approach. 
The main reason for this inaccuracy is that trajectory generation is run using a pre-computing control sequence $\mathcal{U}\triangleq \{u(t_0),\dots, u(t_{N-1}))\} \in \mathcal{D}$, hence any inaccuracy in the learned dynamics would cause the trajectories to go way off the ground truth (GT) due to the nonlinearity of the cart-pole system. This shows the challenging nature of the problem and the proposed approach's efficiency in learning challenging nonlinear dynamics. 
In this context, the superior performance of the proposed approach is due to its alternating optimization approach, especially when compared with hybrid methods such as SPINODE and RM, since estimates of unmeasured states become available when optimizing $\theta$. 
% This feature is unavailable in the competing methods.

 % \vspace{-0.2cm}
\subsection{Electro-mechanical positioning system}
% \vspace{-0.1cm}
% \begin{figure}[htb]
% % \centerline{\includegraphics[trim={0 0 0 0.1cm},clip,width=\myfigurewidth cm]{figure_EMPS.pdf}}
% \centering
% \includegraphics[width=0.8\textwidth]{figure_EMPS.pdf}
% \vspace{-0.3cm}
% \caption{(EMPS)}
% \label{fig:EMPS_result}
% \end{figure}
%

% \setlength{\belowcaptionskip}{-10pt}

%
\begin{figure}
% \centerline{\includegraphics[trim={0 0 0 0.1cm},clip,width=\myfigurewidth cm]{figure_EMPS.pdf}}
\centering
%\vspace{-1.5cm}
%\vspace{-0.3cm}
\includegraphics[width=0.23\textwidth]{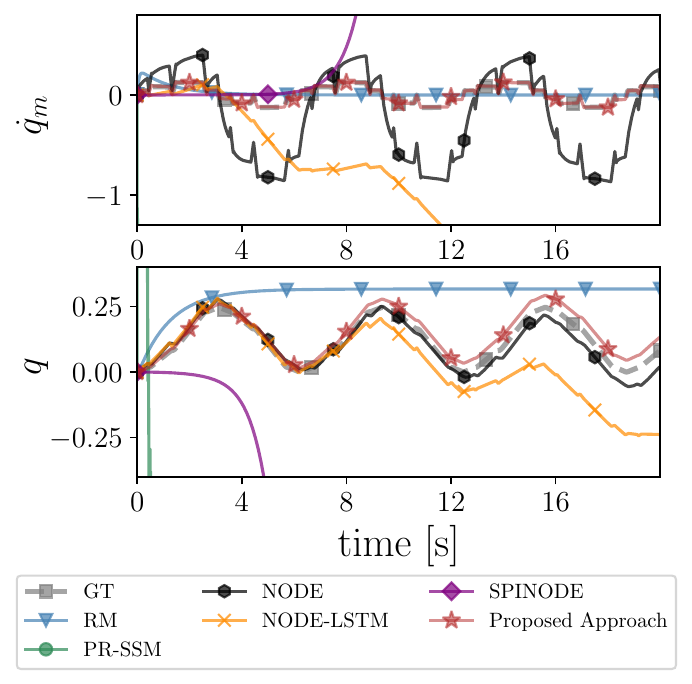}
%\vspace{-0.05cm}
%\captionsetup{font=small}
\caption{Learned state trajectories of EMPS after training RM, PR-SSM, SPINODE, NODE, NODE-LSTM methods and the proposed approach. Results are compared to ground truth ODE system trajectory labeled as GT. The proposed approach can discern the true trajectory for the unmeasured state $\dot{q}_m$.}
\label{fig:EMPS_result}
\end{figure}

%\vspace{-0.3cm}
Here we evaluate the proposed approach on real data from an electro-mechanical positioning system described in \citep{janot2019data}. The training Dataset consists of system's of position, velocity, and control inputs used. The dataset consists of 24801 data points for each state and control input with $dt=10^{-3}s$. Similarly to the HH and cart-pole systems, we train our proposed method using position and control inputs. we replace the velocity's dynamics with a feedforward neural network for all the benchmarks except NODE-LSTM where we used LSTM layers. We used two layers  of 50 and 20 units respectively followed by a $tanh$ activation function for all the benchmarks. 
Table \ref{tab:RMSEs} shows that the proposed approach outperforms the competing algorithms with nRMSE value one to three orders of magnitude smaller than the nRMSEs obtained by the competing methods.  Analyzing the evolution of the latent states depicted 
in Figure~\ref{fig:EMPS_result}, we notice that the proposed approach provides state trajectories that match the ground truth (GT) while SPINODE, PR-SSM, and RM collapse catastrophically.  NODE learns the period of the hidden $\dot{q}_m$ signal but fails the capture its amplitude. The stiffness of $\dot{q}_m$ dynamics plays a role in these results since the sudden jumps shown in Figure~\ref{fig:EMPS_result} are hard to capture. This again demonstrates the robustness of the proposed approach, especially when compared to hybrid physics-based methods such as SPINODE and RM, demonstrating the better performance of the proposed alternating optimization procedure. 

%\vspace{-0.3cm}
\section{Conclusions}
%\vspace{-0.3cm}

We proposed a novel recursive learning mechanism for NODE's to address the challenging task of learning the complex dynamics of ODE systems with partial measurements.
Specifically, we constructed an alternating optimization procedure using Newton's method that sequentially finds optimal system latent states and model parameters. The resulting framework allows for efficient learning of missing ODEs when latent states are identifiable. Different from other competing methods, the proposed approach optimizes model parameters using latent states instead of measured data, leading to superior performance under the partial measurement setting. 
Experiments performed with five complex synthetic systems and one with real data provide evidence that our proposed method is capable of providing adequate solutions in very challenging scenarios.%, attesting our method's superior performance when compared with other state-of-the-art strategies

\bigskip

\bibliography{aaai25}

\appendix

\section{Proof of Theorem 1} \label{app:proof_theo1}
Given a training dataset  $\mathcal{\bm{D}}_N \triangleq \{\bm{d}^{(0)},\dots, \bm{d}^{(N-1)}\}$  with $\bm{d}^{(i)}=\{y(t_i),u(t_i)\}$, determine an optimal solution  $(x^*(t_i), \theta^*(t_i))$  starting from known $(x(t_0),\theta(t_0),y(t_i))$ by solving the following mathematical optimization function: 
%\vspace{-0.25cm}
\begin{equation} 
\begin{aligned}
&\mathcal{L}_N(\Theta_N,X_N)= \\
&\quad\frac{1}{N}\sum_{i=1}^{N}  \lVert x(t_i) - f_o(x(t_{i-1}),u(t_{i-1}),\theta(t_{i-1}))\rVert^{2}_{Q_x^{-1}} \\
&\quad+ \lVert y(t_i) - h(x(t_i)) \rVert^{2}_{R_y^{-1}} + \lVert \theta(t_i) -\theta(t_{i-1})\rVert^{2}_{Q_{\theta}^{-1}} 
\end{aligned}
\label{eq:full_optimization_function_A}
\end{equation} 

Determining optimal solution $(\Theta_N^*,X_N^*)$ using the second order Newton method over the whole data set of size $N$ is computationally expensive. To solve this problem,  we update the optimization function (\ref{eq:full_optimization_function_A}) as follows:
{\small
\begin{equation} 
\begin{aligned}
&\mathcal{L}_{i}(\Theta_i,X_i)=\mathcal{L}_{i-1}(\Theta_{i-1},X_{i-1}) \\ 
&+ \frac{1}{2}\lVert x(t_i) - f_o(x(t_{i-1}),u(t_{i-1}),\theta(t_{i-1})) \rVert^{2}_{Q_x^{-1}}  \\
&+  \frac{1}{2}\lVert y(t_i) - h(x(t_i)) \rVert^{2}_{R_y^{-1}} + \frac{1}{2}\lVert \theta(t_i) -\theta(t_{i-1}) \rVert^{2}_{Q_{\theta}^{-1}}\\
&=\mathcal{L}_{i|i-1}^{x}(\Theta_i,X_i)+ \frac{1}{2}\lVert y(t_i) - h(x(t_i)) \rVert^{2}_{R_y^{-1}}\\
&\scalebox{0.96}{$=\mathcal{L}_{i|i-1}^{\theta}(\Theta_i,X_{i-1})\!\!+\!\!\frac{1}{2} \lVert x(t_i) \!-\! f_o(x(t_{i-1}),u(t_{i-1}),\theta(t_{i-1}))\rVert^{2}_{Q_x^{-1}}$} \\
&\quad+\frac{1}{2} \lVert y(t_i) - h(x(t_i)) \rVert^{2}_{R_y^{-1}}
\end{aligned}
\label{eq:divided_optimization_function_A}
\end{equation}
}
We  start by minimizing the following equation with respect to $x$: 

\begin{equation} 
\begin{aligned}
&\mathcal{L}_{i|i-1}^{x}(\Theta_i,X_i)=\mathcal{L}_{i-1}(\Theta_{i-1},X_{i-1}) \\
&+ \frac{1}{2}\lVert x(t_i) - f_o(x(t_{i-1}),u(t_{i-1}),\theta(t_{i-1})) \rVert^{2}_{Q_x^{-1}}\\
&+ \frac{1}{2} \lVert \theta(t_i) -\theta(t_{i-1}) \rVert^{2}_{Q_{\theta}^{-1}}
\end{aligned}
\label{eq:prediction_optimization_function}
\end{equation}
 $F_{x_{i-1}}=\frac{\partial  f_o(x(t_{i-1}),u(t_{i-1}),\theta(t_{i-1})) }{\partial x(t_{i-1})}$ 

we define the matrix $L_x=[0_{d_x\times d_x},\dots,0_{d_x\times d_x},I_{d_x\times d_x}]$ where $L$ is of dimensions $d_x \times ((i-1)\times d_x) )$ 

by taking the gradient of equation (\ref{eq:prediction_optimization_function}) with respect to $X_i$ we obtain: 
{\small 
\begin{equation} 
\begin{aligned}
&\nabla \mathcal{L}_{i|i-1}^{x}(X_i,\Theta_i)= \\ 
&\left[\begin{matrix} 
    \scalebox{0.9}{$\nabla \mathcal{L}_{i-1}(X_{i-1},\Theta_{i-1}) + L_x^TF_{x_{i-1}}^TQ_x^{-1}\left[x(t_i)-f_o(x(t_{i-1}),\theta(t_{i-1})\right]$}\\
    Q_x^{-1}\left[x(t_{i})-f_o(x(t_{i-1})),\theta(t_{i-1})\right] 
    \end{matrix}\right]
\label{eq:gradient_prediction_x}
\end{aligned}
\end{equation}
}
To minimize (\ref{eq:prediction_optimization_function}), we define the estimate $\hat{X}_{i|i-1}$ of $X_i$ to be the minimizer of  $(\ref{eq:prediction_optimization_function})$. 
by setting $\nabla \mathcal{L}_{i|i-1}^{x}(X_i,\Theta_i)$ to zero we get the following: 
%since $\nabla J_{M-1}=0 $ from the previous update. Therefore results the prediction equations: 

\begin{equation}
    \begin{aligned}
     \hat{X}_{i|i-1}=\left[\begin{matrix} \hat{X}_{i-1} \\
                        f_o(\hat{x}(t_{i-1}),\hat{\theta}(t_{i-1})) \end{matrix}\right]
    \end{aligned}
\end{equation}

with $\hat{x}(t_i|t_{i-1})=f_o(\hat{x}(t_{i-1}),\hat{\theta}(t_{i-1}))$

then, we proceed to minimize the following equation with respect to $\Theta_i$:
{\small 
\begin{equation} 
\mathcal{L}_{i|i-1}^{\theta}(\Theta_i,X_{i-1})= \mathcal{L}_{i-1}(\Theta_{i-1},X_{i-1}) + \frac{1}{2}\lVert \theta(t_i) - \theta(t_{i-1}) \rVert^{2}_{Q_{\theta}^{-1}}  
\label{eq:prediction_optimization_function_theta}
\end{equation}
}
$L_\theta=[0_{d_\theta\times d_\theta},\dots,0_{d_\theta\times d_\theta},I_{d_\theta\times d_\theta}]$ where $L_\theta$ is of dimensions $d_\theta \times ((i-1)\times d_\theta) )$ 

we take the gradient of equation (\ref{eq:prediction_optimization_function_theta}) with respect to $\Theta_i $ we obtain: 
\begin{equation} 
\begin{aligned}
\nabla &\mathcal{L}_{i|i-1}^{\theta}(\Theta_i,X_{i-1})= \\
&\quad\left[\begin{matrix} 
    \nabla \mathcal{L}_{i-1}^{\theta}(\Theta_i,X_{i-1}) - L_\theta^{T}Q_{\theta}^{-1}[\theta(t_{i})-\theta(t_{i-1})] \\
    Q_{\theta}^{-1}[\theta(t_{i})-\theta(t_{i-1})]
    \end{matrix}\right]
\label{eq:gradient_prediction_theta}
\end{aligned}
\end{equation}

To minimize (\ref{eq:prediction_optimization_function_theta}), we define the estimate $\hat{\Theta}_{i|i-1}$ of $\Theta_i$ to be the minimizer of  $(\ref{eq:prediction_optimization_function_theta})$. 
by setting $\nabla \mathcal{L}_{i|i-1}^{\theta}(\Theta_i,X_{i-1})$ to zero we get the following: 
%since $\nabla J_{M-1}=0 $ from the previous update. Therefore results the prediction equations: 

\begin{equation}
    \begin{aligned}
     \hat{\Theta}_{i|i-1}=\left[\begin{matrix} \hat{\Theta}_{i-1} \\
                        \hat{\theta}(t_{i-1}) \end{matrix}\right]
    \end{aligned}
\end{equation}

the second step in the second order Newton method is to calculate the Hessian of $\mathcal{L}_{i|i-1}^{x}(\Theta_i,X_i)$:
\begin{equation}
\begin{aligned}
    \nabla^2 \mathcal{L}&_{i|i-1}^{x}(\Theta_i,X_i)= \\
    &\left[\begin{matrix} \nabla^2 \mathcal{L}_{i-1}^{x}(\Theta_{i-1},X_{i-1}) + O(i) & L_x^{T}F_{x_{i-1}}^{T}Q_x^{-1} \\ 
Q_x^{-1}F_{x_{i-1}}L_x & Q_x^{-1} \end{matrix}\right] 
\end{aligned}
\end{equation}

where 
{\small 
\begin{equation}
\begin{aligned}
O(i)= &-\frac{\partial^2 f(x(t_{i-1}),\theta(t_{i-1})))}{\partial^2 X_{i-1}}Q^{-1}\left[x(t_i)-f(x(t_{i-1}),\theta(t_i))\right] \\ 
&+ L_x^TF_{x_{i-1}}Q_x^{-1}F_{x_{i-1}}L_x 
\end{aligned}
\end{equation}
}
when $X_{i}=\hat{X}_{i|i-1}$ it follows that $O(i)=L_x^TF_{x_{i-1}}^TQ_x^{-1}F_{x_{i-1}}L_x$.
using Lemma B.3 in \citep{humpherys2012fresh}, the lower block $ P_{x_i}^{-}$ of $ \nabla^{2} \mathcal{L}_{i|i-1}^{x}(\Theta_i,X_i)$  is calculated as follows: 

\begin{equation}
P_{x_i}^{-}=Q_x^{-1} + F_{x_{i-1}}^{T}P_{x_{i-1}}F_{x_{i-1}}
\label{eq:prediction_covariance}
\end{equation}

we continue to minimize 
\begin{equation}
    \mathcal{L}_{i}^{x}(X_i,\Theta_i)=\mathcal{L}_{i|i-1}^{x}(X_i,\Theta_i) + \frac{1}{2}\lVert y(t_i)- h(x(t_{i}))\rVert^2_{R_y^{-1}}
\label{eq:update_optimization_function}
\end{equation}

we denote $$\mathcal{H}_i=\left[0,0,\dots, \frac{\partial h(x(t_i)}{\partial x(t_i)} \right]. $$
and by $H_i=\frac{\partial h(x(t_i))}{\partial x(t_i)}$
by taking the gradient of equation (\ref{eq:update_optimization_function}) we obtain: 
$$\nabla \mathcal{L}_{i}^{x}(X_i,\Theta_i)=\nabla \mathcal{L}_{i|i-1}^{x}(X_i,\Theta_i) +\mathcal{H}_i R_y^{-1}(y(t_i)- h(x(t_{i}))) $$

The hessian of (\ref{eq:update_optimization_function}) becomes:

\begin{equation}
\begin{aligned}
    \nabla^2  \mathcal{L}_{i}^{x}(X_i,\Theta_i)&=\nabla^2 \mathcal{L}_{i|i-1}^{x}(X_i,\Theta_i) \\ &+\frac{\partial^2 \mathcal{H}_i}{\partial^2 X_{i}} R_y^{-1}(y(t_i)- h(x(t_i))) + \mathcal{H}_iR_y^{-1}\mathcal{H}_i
\end{aligned}
\end{equation}

setting $X_{i}=\hat{X}_{i|i-1} \implies \nabla \mathcal{L}_{i}^{x}(X_i,\Theta_i)=0$ therefore:

\begin{equation} 
\nabla \mathcal{L}_{i}^{x}(X_i,\Theta_i)=\left[\begin{matrix} 
    0 \\
    H_{_i}R_y^{-1}(y_i-h_{i}(\hat{x}(t_i|t_{i-1})) 
    \end{matrix}\right]
\label{eq:gradient_update}
\end{equation}

The hessian hence becomes:
\begin{equation}
   \nabla^2 \mathcal{L}_{i}^{x}(X_i,\Theta_i)=\nabla^2 \mathcal{L}_{i|i-1}^{x}(X_i,\Theta_i)+ \mathcal{H}_{i}R_y^{-1}\mathcal{H}_{i}
\end{equation}

then according to the Newton method, we can update our estimate of $x_{i}$ as follows:
\begin{equation}
\hat{X}_{i}=\hat{X}_{i|i-1} -\left(\nabla^2 \mathcal{L}_{i}^{x}\right)^{-1}\nabla \mathcal{L}_{i}^{x}
\label{eq:newton_equation}
\end{equation} 
let $P_{x}$ be the bottom right block of $\left(\nabla^2 \mathcal{L}_{i}^{x}\right)^{-1}$, therefore 
\begin{equation} 
\begin{aligned}
P_{x_i}=&P_{x_i}^{-} + P_{x_i}^{-}H_{i}\left(R_y-H_{i}P_{x_i}^{-}H_{i}^{T}\right)H_{i}P_{x_i}^{-}\\
=&\left((P_{x_i}^{-})^{-1} + H_{i}^{T}R_y^{-1}H_{i}\right)^{-1}
\end{aligned}
\end{equation} 

by taking the bottom row of the newton equation (\ref{eq:newton_equation}) we get
\begin{equation}
    \begin{aligned}
        \hat{x}(t_i)&=\hat{x}(t_i|t_{i-1}) -\left[\left(\nabla^2 \mathcal{L}_{i}^{x}(\Theta_i,X_i)\right)^{-1}\right]_{i,:}\nabla \mathcal{L}_{i}^{x}(\Theta_i,X_i) \\
        &=\hat{x}(t_i|t_{i-1}) - P_{x_i}H_iR_y^{-1}\left(h(\hat{x}({t_i|t_{i-1}}))-y(t_i)\right) \\
        &=\hat{x}(t_i|t_{i-1}) - K_i^{x}\left(h(\hat{x}(t_i|t_{i-1}))-y(t_i)\right)
    \end{aligned}
\end{equation}
where $row_i$ corresponds to the $i^{th}$ row of matrix $ \left(\nabla^2 \mathcal{L}_{i}^{x}(\Theta_i,X_i)\right)^{-1}$ and 
\begin{equation}
    \begin{aligned}
    K_{i}^{x}&=P_{x_i}H_iR_y^{-1}\\
    &=P_{x_i}^{-}H_i^T\left(H_iP_{x_i}^{-}H_i^T + R_y\right)^{-1}
    \end{aligned}
    \label{eq:kalman_gain}
\end{equation}

In a similar fashion we proceed to minimize 
\begin{equation}
\begin{aligned}
\mathcal{L}_i^{\theta}(\Theta_i,X_i)&=\mathcal{L}_{i|i-1}^{\theta}(\Theta_i,X_{i-1})\\ & + \frac{1}{2}\lVert x(t_i) - f_o(x(t_{i-1}),u(t_i),\theta(t_i))  \rVert^{2}_{Q_x^{-1}}\\
&+ \frac{1}{2}\lVert y(t_i) - h(x(t_i)) \rVert^{2}_{R_y^{-1}} 
\label{eq:L_theta_A}
\end{aligned}
\end{equation}

\begin{equation} 
\begin{aligned}
&\nabla \mathcal{L}_{i}^{\theta}(X_i,\Theta_i)= \\
&\left[\begin{matrix} 
    \scalebox{0.9}{$\nabla \mathcal{L}_{i|i-1}(X_{i-1},\Theta_{i-1}) + L_{\theta}^TF_{\theta_{i-1}}^TQ_x^{-1}\left[x_{i}-f(x(t_{i-1}),\theta(t_{i-1})\right]$}\\
     Q_{\theta}^{-1}[\theta(t_{i})-\theta(t_{i-1})]
    \end{matrix}\right]
\label{eq:gradient_theta_1}
\end{aligned}
\end{equation}

where $F_{\theta_{i-1}}=\frac{\partial  f_o(x(t_{i-1}),u(t_{i-1}),\theta(t_{i-1})) }{\partial \theta(t_{i-1})}$ 

at $\Theta_{i}=\hat{\Theta}_{i|i-1}$ , 
\begin{equation} 
\nabla \mathcal{L}_{i}^{\theta}(X_i,\Theta_i)=\left[\begin{matrix} 
    L_{\theta}^TQ_x^{-1}\left[x_{i}-f(x(t_{i-1}),\theta(t_{i-1})\right] \\
    0
    \end{matrix}\right]
\label{eq:gradient_theta_2}
\end{equation}

Similarly, the hessian of (\ref{eq:update_optimization_function}) is:

{\small 
\begin{equation}
\begin{aligned}
\nabla^2 \mathcal{L}_{i}^{\theta}(\Theta_i,X_i)\!=\!\left[\begin{matrix} \nabla^2 \mathcal{L}_{i-1}^{\theta}(\Theta_{i-1},X_{i-1}) + Z(i) & \!\!L_{\theta}^{T}F_{\theta_{i-1}}^{T}Q_{\theta}^{-1} \\ 
Q_{\theta}^{-1}F_{\theta_{i-1}}L_{\theta} & Q_{\theta}^{-1} \end{matrix}\right]
\end{aligned}
\end{equation}
}
where $$\!Z(i)\!=\!\frac{\partial^2  f_o(x(t_{i-1}),u(t_{i-1}),\theta(t_{i-1})) }{\partial^2 \theta(t_{i-1})}Q_x^{-1}\left[x_{i}\!-\!f(x(t_{i-1}))\right]$$ 
$$\hspace{-1cm}+ L_{\theta}^{T}F_{\theta_{i-1}}Q_x^{-1}F_{\theta_{i-1}}L_{\theta} + L_{\theta}^TQ_{\theta}^{-1}L_{\theta}$$
by ignoring second order terms we obtain:
$Z(i)\!=\!L_{\theta}^{T}F_{\theta_{i-1}}Q_x^{-1}F_{\theta_{i-1}}L_{\theta} + L_{\theta}^TQ_{\theta}^{-1}L_{\theta}$

Then according to the newton second order method, we can update our estimate of $\Theta_{i}$ as follows:
\begin{equation}
\hat{\Theta}_i=\hat{\Theta}_{i|i-1} -\left(\nabla^2 \mathcal{L}_{i}^{\theta}\right)^{-1}\nabla \mathcal{L}_{i}^{\theta}
\label{eq:newton_equation_theta}
\end{equation} 
let $P_{\theta}$ be the bottom right block of $\left(\nabla^2 J_{i}^{\theta}\right)^{-1}$, therefore 
{\small
\begin{equation} 
\begin{aligned}
\!P_{\theta_i}\!\!=&Q_{\theta}\!+\!\!\left[\!P_{\theta_{i-1}} \!\!\!-\!P_{\theta_{i-1}}F_{\theta_{i-1}}^{T}\!\!\left(\!Q_x\!+\!F_{\theta_{i-1}}\!P_{\theta_{i-1}}\!F_{\theta_{i-1}}^{T}\!\right)\!F_{\theta_{i-1}}\!P_{\theta_{i-1}}\right]\\
=&Q_{\theta}\!+\!P_{\theta}^{-}
\end{aligned}
\end{equation} 
}
where $$P_{\theta_i}^{-}\!=\!P_{\theta_{i-1}} \!-P_{\theta_{i-1}}F_{\theta_{i-1}}^{T}\!\left(\!Q_x\!+\!F_{\theta_{i-1}}P_{\theta_{i-1}}F_{\theta_{i-1}}^{T}\!\right)F_{\theta_{i-1}}P_{\theta_{i-1}}$$ 

by taking the bottom of the newton equation (\ref{eq:newton_equation}) we get
\begin{equation}
    \begin{aligned}
        \hat{\theta}(t_i)&\!=\!\hat{\theta}(t_i|t_{i-1}) \!-\!\!\left[\!\left(\!\nabla^2 \!D\mathcal{L}_{i}^{\theta}(\Theta_i,X_{i-1})\right)^{-1}\right]_{i,:}\!\!\!\!\!\!\nabla \mathcal{L}_{i}^{\theta}(\Theta_i,X_{i-1})\\
        &=\hat{\theta}(t_i|t_{i-1}) +K_{i}^{\theta}\left(\hat{x}(t_i)-f(\hat{x}(t_i),\hat{\theta}(t_i)\!\right) \\
    \end{aligned}
\end{equation}

with
\begin{equation}
    \begin{aligned}
    K_{i}^{\theta}&= P_{\theta_i}^{-}F_{\theta_{i-1}}^{T}
    \end{aligned}
    \label{eq:kalman_gain_theta}
\end{equation}

\section{Vanishing gradients}\label{app:vanishingGrad}

Under the joint optimization scenario according to \citep{humpherys2012fresh}, we define the joint state $z(t_{i-1})=[\theta(t_i),x(t_i)] $ and $Z(t_i)=[\Theta_{i-1},X_{i-1}]$. By following the same derivation as \citep{humpherys2012fresh} we get the following update equation :
\begin{equation}
\hat{Z}_i=\hat{Z}_{i|i-1} -\left(\nabla^2 \mathcal{L}_{i}(Z)\right)^{-1}\nabla \mathcal{L}_{i}(Z)
\label{eq:newton_equation_Z}
\end{equation}

the upper rows of \eqref{eq:newton_equation_Z} correspond to the parameter update which takes the following form \citep{wan2001dual}: 

\begin{equation}
    \begin{aligned}
        \hat{\theta}(t_i)&=\hat{\theta}(t_i|t_{i-1}) + K_{i}^{\theta}\left(y(t_i)-h(\hat{x}(t_i))\right) 
    \end{aligned}
    \label{eq:theta_update_joint}
\end{equation}

with
\begin{equation}
    \begin{aligned}
    K_{i}^{\theta}&= P_{\theta_i}^{-}F_{\theta_{i-1}}^{T}H_{i}^T\left[H_{i}F_{\theta_{i-1}}P_{\theta_i}^{-}F_{\theta_{i-1}}^{T}H_{i}^T + Q_x\right]^{-1}
    \end{aligned}
    \label{eq:kalman_gain_theta_joint}
\end{equation}

in the case of partial measurements, \eqref{eq:kalman_gain_theta_joint} vanishes since the product $F_{\theta_{i-1}}^{T}H_{i}^T$ becomes a zero matrix.

To see this let's use the use the example of a 3-state ODE with state vector $x=[x_1,x_2,x_3]$, where the last ODE is not measured and replaced by a neural network $a(\cdot, \theta)$, parameterized by $\theta$. Since $x_3$ is not measured $h(x)=[x_1,x_2]$.
The ODE system can be then written as:
 \begin{equation}
     \begin{aligned}
     \dot{x}=f(x,\theta)+ \epsilon
     \end{aligned}
 \end{equation}
 where 
 
 \begin{equation}
     \begin{aligned}
     \dot{x}_1&=f_1(x) + \epsilon_1\\
     \dot{x}_2&=f_2(x) + \epsilon_2\\
     \dot{x}_3&=a(x,\theta)+ \epsilon_3
     \end{aligned}
 \end{equation}
 
 The gradients of $f$ and $h$ are presented below:  
 \begin{equation}
    F_{\theta_{i-1}}=\left[\begin{matrix} 0 \dots  0 \\ 
 0 \dots  0 \\
 \frac{\partial  a(x(t_{i-1}),\theta(t_{i-1})) }{\partial \theta(t_{i-1})}  \end{matrix}\right] 
\end{equation}
and 
 \begin{equation}
    H_i=\left[\begin{matrix} 1 & 0 & 0 \\ 
   0 & 1 & 0 \\ \end{matrix}\right] 
\end{equation}

 Therefore, multiplying these two matrices in \eqref{eq:kalman_gain_theta_joint} results in: 
  \begin{equation}
  \begin{aligned}
    F_{\theta_{i-1}}^{T}H_i^{T}=\left[\begin{matrix} 0 & 0 \\ 
  \vdots  & \vdots \\
  0 & 0 \end{matrix}\right] 
 \end{aligned}
 \label{eq:vanishing_grad}
\end{equation}

by replacing (\ref{eq:vanishing_grad}) in equation (\ref{eq:kalman_gain_theta_joint}), the latter vanishes too, and the update of (\ref{eq:theta_update_joint}) fails.

\section{Models and further experiments}

\subsection{Yeast Glycolysis Model}
Yeast glycolysis Model is a metabolic network that explains the process of breaking down glucose to extract energy in the cells. This model has been tackled by similar works in the field  \citep{kaheman2020sindy}, \citep{mangan2016yeast}, and \citep{Schmidt2011AutomatedRA}. It has seven states: $x = \begin{bmatrix} x_1  & x_2 & x_3 & x_4 & x_5  & x_6  & x_7  \end{bmatrix}^T$, and ODEs for these states are given from Eq (\ref{eq:yeast_glycolysis}), \citep{mangan2016yeast}.

\begin{equation}
\begin{aligned}
\dot{x}_1 &= c_1 + \frac{ c_2x_1x_6}{1 + c_3x_6^{4}},  \\ 
\dot{x}_2 &= \frac{d_1x_1x_6}{1 + d_2x_6^{4}} + d_3x_2 - d_4x_2x_7, \\
\dot{x}_3 &= e_1x_2 + e_2x_3 + e_3x_2x_7 + e_4x_3x_6, \\
\dot{x}_4 &= f_1x_3 + f_2x_4 + f_3x_5 + f_4x_3x_6 + f_5x_4x_7, \\
\dot{x}_5 &= g_1x_4 + g_2x_5, \\
\dot{x}_6 &= h_3x_3 + h_5x_6 + h_4x_3x_6 + \frac{h_1x_1x_6}{1 + h_2x_6^{4}}, \\
\dot{x}_7 &= j_1x_2 + j_2x_2x_7 + j_3x_4x_7 
\end{aligned}
\label{eq:yeast_glycolysis}
\end{equation}

\begin{figure}[htb]
% \hspace{0.1cm}
\centerline{\includegraphics[trim={0 0 0 0.1cm},clip,width=9 cm]{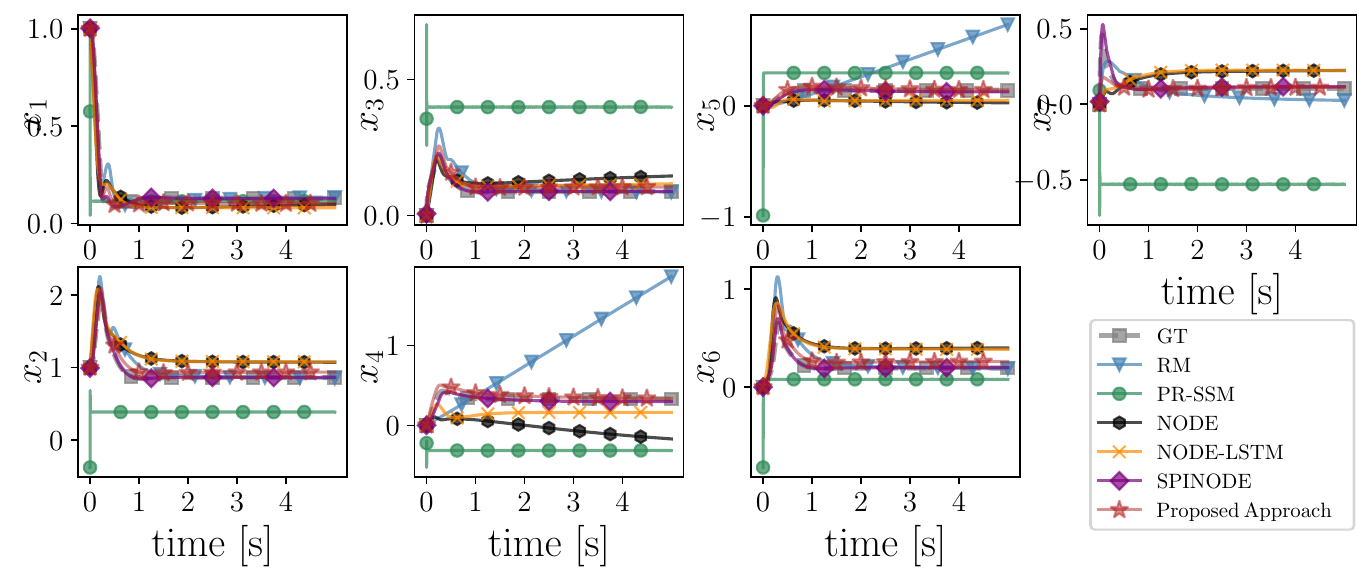}} 
% \vspace{-0.3cm}
\caption{Learned state trajectories of yeast glycolysis model after training with the RM, PR-SSM, NODE, NODE-LSTM,SPINODE methods and the proposed approach. Results are compared to ground truth ODE system trajectory labeled as GT. We showed that the proposed approach  is capable of discerning the true trajectory for the unmeasured state $x_4$.}
\label{fig:yeast_result}
\end{figure}
In this scenario, we assumed the dynamics of state $\dot{x}_4$ in \eqref{eq:yeast_glycolysis} to be unknown and $x_4$ unmeasured. We used the proposed approach and competing algorithms to learn NN-based dynamics. Specifically, we replaced $\dot{x}_4$ with a two-layer neural network with $tanh$ activation function on each layer. 
% We don’t measure yeast glycolysis's variable $x_4$.
We generate a dataset with 5000 time steps, $dt=10^{-3}s$ and by collecting measurements $\mathcal{D} \triangleq \{y(t_0),\dots,y(t_{N-1})\}$. 
For the proposed approach the hyper-parameters were set to $P_{x_0}=10^{-2}I_{d_x}, P_{\theta_0}=10^{2}I_{d_{\theta}}$ $R_y =10^{-10}I_{d_y}, Q_x=10^{-5}I_{d_x}$ and $ Q_{\theta}=10^{-2}I_{d_{\theta}}$. At the beginning of each epoch, we solve problem (\ref{eq:optim_x0}) of the Appendix \ref{app:x0_appendix} to get the initial condition.  Final optimal parameters $\hat{\theta}(t_N)$ and initial condition $\hat{x}(t_0)$ are saved and collected at the end of training.  

We assess the performance of our proposed approach by setting the model parameters to $\hat{\theta}(t_N)$ and perform integration following the model described in \eqref{eq:trajectory_generation} starting from initial condition $\hat{x}(t_0)$.
In Figure \ref{fig:yeast_result}, we demonstrate the ability of the proposed approach to learning the underlying dynamics of the partially measured system compared to RM, PR-SSM and NODE methods.  
Table \ref{tab:RMSEs}, shows that the proposed approach clearly outperforms the competing algorithms with nRMSE 99.3\% , 99.1\% and 90.4\% smaller than the nRMSEs obtained by PR-SMM, RM, and NODE respectively. Analyzing the evolution of the latent states depicted in Figure~\ref{fig:yeast_result}, we notice that the proposed approach provides state trajectories that match the ground truth (GT). PR-SSM fails to capture the dynamics of the system, while RM and NODE present acceptable trajectories of most of the states except for the unmeasured dynamics of $\dot{x}_4$, and the measured dynamics of $\dot{x}_5$.

Moreover, both RM  and NODE state trajectories are much more inaccurate than the one provided by the proposed approach.  This shows the challenging nature of the problem and the proposed approach's efficiency in learning challenging nonlinear dynamics using estimates of the unmeasured states, which is unavailable to the other methods.

\subsection{Hodgkin-Huxley Neuron Model} \label{HH_appendix}
For the HH model, we refer to the \citep{Hodgkin_1952} and use the following ODE system. The ODE system has four states: $V_m$, $n_{gate}$, $m_{gate}$, and $h_{gate}$. $I_e$ is the external current input, which is set to 10 if the neuron is firing, and 0 otherwise. For all models, we simulate the dynamics of the HH model with a time step of 0.01 ms and integrate using Euler integration. 

% \begin{align}
%     \dot{V}_m &= I_e - 36 n_{gate}^4 (V_m + 77) \nonumber \\
%               &\quad - 120 m_{gate}^3 h_{gate} (V_m - 50) - 0.3 (V_m + 54.4)  \label{v_m_states} \\
%     \dot{n}_{gate} &= 0.01 (V_m + 55) \!\left[ 1 \!- \!\exp \left( -\frac{V_m + 55}{10} \right) \right]^{-1} \!\!\!\!\!(1 - n_{gate}) \nonumber \\
%                    &\quad - 0.125 \exp \left( -\frac{V_m + 65}{80} \right) n_{gate} \label{eq:ngate} \\ 
%     \dot{m}_{gate} &= 0.1 (V_m + 40) \!\left[ 1 \!- \!\exp \left( -\frac{V_m + 40}{10} \right) \right]^{-1} \!\!\!\!\!(1 - m_{gate}) \nonumber \\
%                    &\quad - 4 \exp \left( -\frac{V_m + 65}{18} \right) m_{gate} \label{eq:mgate} \\
%     \dot{h}_{gate} &= 0.07 \exp \left( -\frac{V_m + 65}{20} \right) (1 - h_{gate}) \nonumber \\
%                    &\quad - \left[ 1 + \exp \left( -\frac{V_m + 35}{10} \right) \right]^{-1} h_{gate} \label{eq:dq}
% \end{align}

{\small
\begin{align}
    \dot{V}_m &= I_e - 36 n_{gate}^4 (V_m + 77) \nonumber \\
              &\quad - 120 m_{gate}^3 h_{gate} (V_m - 50) \nonumber \\
              &\quad - 0.3 (V_m + 54.4)  \label{v_m_states} \\
    \dot{n}_{gate} &= 0.01 (V_m + 55) (1 - \exp( \scalebox{0.9}{$- (V_m + 55) / 10$}))^{-1} (1 - n_{gate}) \nonumber \\
                   &\quad - 0.125 \exp(\scalebox{0.9}{$- (V_m + 65) / 80$}) n_{gate} \label{eq:ngate} \\ 
    \dot{m}_{gate} &= 0.1 (V_m + 40) (1 - \exp(\scalebox{0.9}{$- (V_m + 40) / 10$}))^{-1} (1 - m_{gate}) \nonumber \\
                   &\quad - 4 \exp(\scalebox{0.9}{$- (V_m + 65) / 18$}) m_{gate} \label{eq:mgate} \\
    \dot{h}_{gate} &= 0.07 \exp(\scalebox{0.9}{$- (V_m + 65) / 20$}) (1 - h_{gate}) \nonumber \\
                   &\quad - (1 + \exp(\scalebox{0.9}{$- (V_m + 35) / 10$}))^{-1} h_{gate} \label{eq:dq}
\end{align}
}

\subsection{Retinal Circulation Model} \label{retina_appendix}
\begin{figure}[htb]
\centerline{\includegraphics[trim={0 0 0 0.1cm},clip,width=0.5\textwidth]{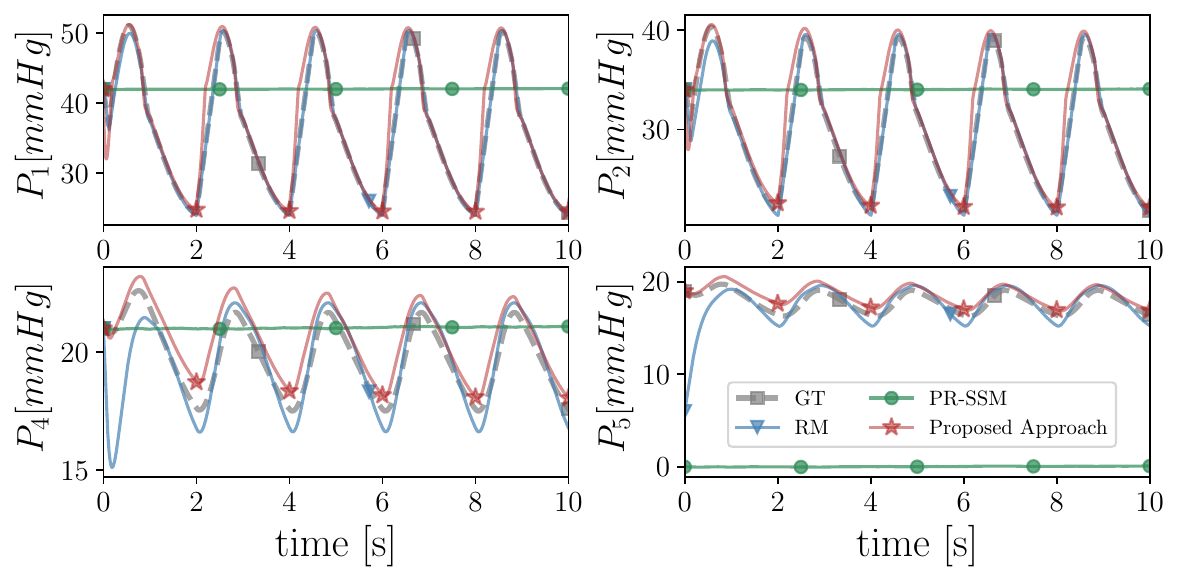}}
\caption{Estimated state trajectories of Retinal circulation model after training RM, PR-SSM methods and the proposed approach. Results are compared to ground truth ODE system trajectory labeled as GT. We showed that the proposed approach is capable of discerning the true trajectory for the unmeasured state $P_{5}$.}
\label{fig:retina_result}
\end{figure}

The retinal circulation model describes the internal pressures of five compartments in the retina \citep{guidoboni2014intraocular}. The model has four states: $P_{1}$, $P_{2}$, $P_{4}$, and $P_{5}$. The relation between these states is summarized in Eqs. (\ref{eq:p1})-(\ref{eq:p5}). In our experiments, we don't measure $P_5$ and set $y= (P_1, P_2, P_4)$ then train RM, PR-SSM, and the proposed approach to approximate the ODE trajectories. In Fig. \ref{fig:retina_result}, we visualize the state trajectories for all states and demonstrate that the proposed approach outperforms PR-SSM and RM at estimating state trajectories, our proposed mechanism successfully captures the dynamics of the unmeasured $P_5$ state.

% \begin{align}
%     \dot{P_1} = &  \, \mathcal F_1(P_1,P_2;P_{in}) \label{eq:p1} \\
%     \dot{P_2} = &  \, \mathcal F_2(P_1,P_2,P_4)  \label{eq:p2} \\
%     \dot{P_4} = &  \, \mathcal F_4(P_2,P_4,P_5)  \label{eq:p4} \\
%     \dot{P_5} = &  \, \mathcal F_5(P_4,P_5;P_{out}) \label{eq:p5}
% \end{align}
{\small
\begin{align}
    \dot{P_1} = &  \, \frac{P_{in} - P_1}{C_1(R_{\text{in}} \!+\! R_{1a})}         - \frac{P_1 - P_2}{C_1(R_{1b} \!+ \!R_{1c} \!+\! R_{1d} \!+\! R_{2a})} \label{eq:p1} \\
    \dot{P_2} = &  \, \frac{P_1 - P_2}{C_2(R_{1b} \!+\! R_{1c} \!+\! R_{1d} \!+\! R_{2a})} - \frac{P_2 - P_4}{C_2(R_{2b} \!+\! R_{3a} \!+\! R_{3b} \!+\! R_{4a})}  \label{eq:p2} \\
    \dot{P_4} = &  \, \frac{P_2 - P_4}{C_4(R_{2b} \!+\! R_{3a} \!+\! R_{3b} \!+\! R_{4a})} - \frac{P_4 - P_5}{C_4(R_{4b} \!+\! R_{5a} \!+\! R_{5b} \!+\! R_{5c})}  \label{eq:p4} \\
    \dot{P_5} = &  \, \frac{P_4 - P_5}{C_5(R_{4b} \!+\! R_{5a} \!+\! R_{5b} \!+\! R_{5c})} - \frac{P_5 - P_{out}}{C_5(R_{5d} \!+\! R_{out})} \label{eq:p5}
\end{align}
}
$R_{in}, R_{1a}, R_{1b}, R_{2a}, R_{2b}, R_{3a}, R_{3b}, R_{5c}, R_{5d}$,  and $R_{out}$ are fixed resistances. $R_{4a}, R_{4b}, R_{5a}$, and $R_{5b}$ depend on states. $C_{1-5}$ are the constant capacitance values. $P_{in}$ is time-varying input, and $P_{out}$ is constant output which is set to 14.

The numerical results for the retinal circulation model experiments are summarized in Table \ref{tab:RMSEs} and visual results are presented in Figure \ref{fig:retina_result}. Both results show that the proposed approach clearly outperforms the competing algorithms with nRMSE value that is 94.3\%  and 28.8\% smaller than the nRMSEs obtained by PR-SMM and RM, respectively. Analyzing the evolution of the latent states depicted in Figure~\ref{fig:retina_result}, we notice that the proposed approach provides state trajectories that match the ground truth (GT) more closely when compared with PR-SSM and RM. Similar to PR-SSM results for the HH, HO, and EMPS models, PR-SSM again presents very poor state trajectories indicating that the model was not capable of learning the underlying ODE function accurately. nRMSE value for the RM is comparable to the proposed approach, however, the lower right panel of the Fig. \ref{fig:retina_result} pinpoints that the proposed approach excels in learning the unmeasured state $P_5$.

\subsection{Harmonic Oscillator} \label{ho_appendix}
The harmonic oscillator has two states, representing position and velocity, where $z$ is the position and $\dot{z}$ is the velocity: $x = \begin{bmatrix} z  &  \dot{z} \end{bmatrix}^T$. \(u\) is the input vector which is set to zero for a free harmonic oscillator: \(u = \begin{bmatrix} 0 \end{bmatrix}\).
\(\omega\) is the unknown angular frequency. State equations can be written in matrix form as follows:
   
\[ \dot{x} = \begin{bmatrix} 0 & 1 \\ -\omega^2 & 0 \end{bmatrix} x +  \begin{bmatrix} 0 \\   1 \end{bmatrix} u \]

Here, the matrix \(\begin{bmatrix} 0 & 1 \\ -\omega^2 & 0 \end{bmatrix}\) is the state transition matrix and it represents the system's dynamics. Throughout the experiments, we only measured position state $x_1$, hence, our output equation is: $ y = \begin{bmatrix} 1 & 0 \end{bmatrix} x$. We simulated the ODE system using Euler integration and we used a time step of 1 ms.

\begin{figure}[htb]
% \hspace{0.1cm}
\centerline{\includegraphics[trim={0 0 0 0.1cm},clip,width=0.5\textwidth]{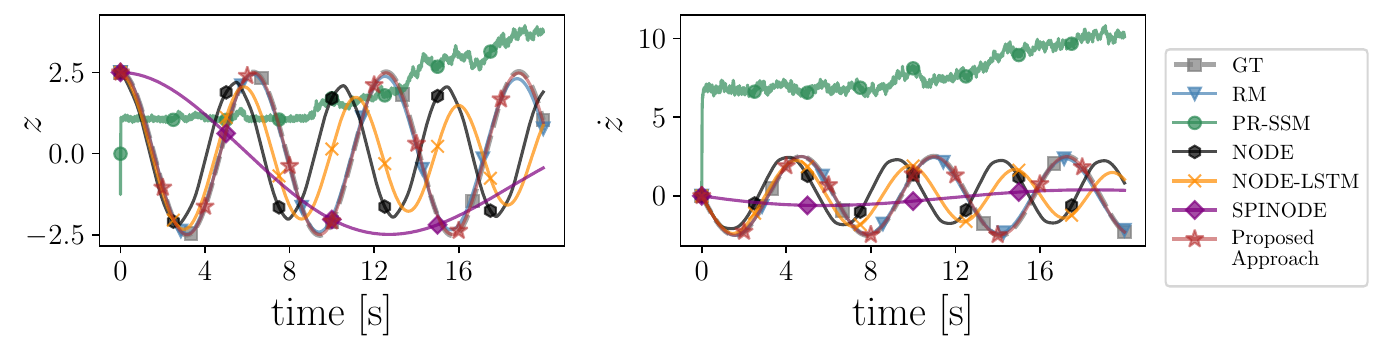}} 
\caption{Estimated state trajectories of harmonic oscillator after training RM, PR-SSM, NODE, NODE-LSTM, SPINODE methods and the proposed approach. Results are compared to ground truth ODE system trajectory labeled as GT. We showed that the proposed approach is capable of discerning the true trajectory for the unmeasured state $\dot{z}$.}
\label{fig:ho_result}
\end{figure}

The numerical results for the HO experiments are summarized in Table \ref{tab:RMSEs} while the corresponding visual results can be found in Figure \ref{fig:ho_result}. Both results clearly demonstrate our approach's superior performance against PR-SSM and NODE, and Table \ref{tab:RMSEs} shows a modest advantage over RM. Our approach achieved notably smaller nRMSE. Indeed, the nRMSE achieved using the proposed approach is 99\%, 78\% and 98\% smaller than the nRMSEs obtained by PR-SMM, RM and NODE respectively. Analyzing the evolution of the states in Figure~\ref{fig:ho_result}, we notice that PR-SSM failed to learn the underlying ODE function accurately and that NODE failed to learn the period of the signal.

\begin{figure}[htb]
% \hspace{0.1cm}
\centerline{\includegraphics[trim={0 0 0 0.1cm},clip, width=0.5\textwidth]{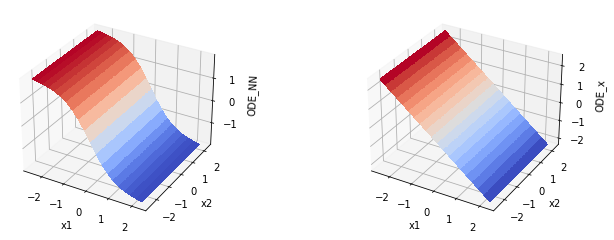}} 
\caption{Learned ODE\_NN (left) and true ODE\_x (right) vector fields of the Harmonic Oscillator model after training with the proposed approach. We showed that the proposed approach is capable of learning the true vector field.}
\label{fig:ho_vector_field}
\end{figure}

Figure~\ref{fig:ho_vector_field} depicts the learned ODE vector field (left) and true vector field (right). We can observe that the proposed approach was capable of learning reasonably well the true ODE function.

\subsection{Cart-Pole System} \label{CP_appendix}
The cart-pole system is a classic problem in control theory and it models the movement of a cart along an axis, and this cart has a pole attached to it and this pole can pivot freely. States of the ODE system can be defined as: $x = \begin{bmatrix} z  & \dot{z} & \phi  & \dot{\phi} \end{bmatrix}^T$. We linearize the system at $x = \begin{bmatrix} z  & \dot{z} & \phi  & \dot{\phi} \end{bmatrix}^T = \begin{bmatrix} 0 & 0 & 0 & 0 \end{bmatrix}^T$ and use LQR controller \citep{prasad2011optimal} to calculate the input $u$ that swings the pendulum and balances it in
the inverted position in the middle of the track. 
% After linearization, the matrices \( \mathbf{A} \) and \( \mathbf{B} \) for the cart-pole system can be written as follows:

\begin{align}
\!\!\dot{z} = & \dot{z}  \label{eq:cartpole_state1} \\
\!\!\ddot{z} = & \frac{-m \cdot l \cdot \sin(\phi) \cdot \dot{\phi} + u + m \cdot g \cdot \cos(\phi) \cdot \sin(\phi)}{M + m - m \cdot \cos(\phi)^2} \label{eq:cartpole_state2} \\
\!\!\dot{\phi} = & \dot{\phi} \label{eq:cartpole_state3} \\
\!\!\ddot{\phi} \!= & \frac{\scalebox{0.87}{$\!\!-m \!\cdot \!l \!\cdot \!\cos(\phi)\! \cdot \!\sin(\phi)\! \cdot \!\dot{\phi}^2 \!+\! u \!\cdot\! \cos(\phi) \!+\! m \!\cdot \!g \!\cdot \!\sin(\phi) \!+\! M\! \!\cdot \!g\! \cdot \sin(\phi)$}}{l \cdot (M + m - m \cdot \cos(\phi)^2)}  \label{eq:cartpole_state4}  
 \end{align}

% \begin{equation}
% \mathbf{\dot{x}} = \begin{bmatrix} \dot{x}  \\  \ddot{x} \\  \dot{\theta}  \\  \ddot{\theta} \end{bmatrix} =  \begin{bmatrix} \dot{x} \\ \frac{-m \cdot l \cdot \sin(\theta) \cdot \dot{\theta} + u + m \cdot g \cdot \cos(\theta) \cdot \sin(\theta)}{M + m - m \cdot \cos(\theta)^2} \\  \dot{\theta} \\ \frac{-m \cdot l \cdot \cos(\theta) \cdot \sin(\theta) \cdot \dot{\theta}^2 + u \cdot \cos(\theta) + m \cdot g \cdot \sin(\theta) + M \cdot g \cdot \sin(\theta)}{l \cdot (M + m - m \cdot \cos(\theta)^2)} \end{bmatrix} 
% \label{eq:cartpole_state1} \end{equation} 

% \begin{equation}
% \mathbf{A} = \begin{bmatrix}
% 0 & 1 & 0 & 0 \\
% 0 & 0 & \frac{mg}{M}  & 0 \\
% 0 & 0 & 0 & 1 \\
% 0 & 0 & \frac{(m+M)g}{Ml} & 0
% \end{bmatrix} \quad 
% \mathbf{B} = \begin{bmatrix} 0 \\ \frac{1}{M} \\ 0 \\ \frac{1}{Ml} \end{bmatrix}
% \end{equation} \label{eq:cartpole_statespace}

In these matrices: \( M \) is the mass of the cart, \( m \) is the mass of the pole, \( l \) is the length from the cart's center to the pole's center of mass, \( l_c \) is the length from the cart's center to the pivot point, and \( g \) is the acceleration due to gravity.

% \subsection{Electro-Mechanical Positioning System} \label{EMPS_appendix}

% \begin{align}
% \dot{q}_m(t) = & \, \frac{q_{fm}(t) \, \text{-} \,  q_{fm}(t-1)}{dt} \label{eq:emps1}  \\ 
% q_{fm}(t) = & \, q_{m}(t)/2 + q_{m}(t-1)  \label{eq:emps2}   
% \end{align}
% where $dt$ is the sampling time (1ms)

\section{Initial Condition Reconstruction During Training }\label{app:x0_appendix}
Given a model $\theta$ and a dataset $\mathcal{D} \triangleq \{u(t_0),y(t_0),\dots, u(t_{N-1}),y(t_{N-1})\}$ , training the model  requires determining
an appropriate initial state $x(t_0)$ at the beginning of each epoch. A way to get $x(t_0)$ is to solve the following state-reconstruction problem: 
\begin{equation}
    \min_{x(t_0|t_{-1})} \quad \lVert y(t_0) - h(\hat{x}(t_0)) \rVert^{2}_{R_y^{-1}} 
    \label{eq:optim_x0}
\end{equation}
%where $\hat{x}(t_{0})=\hat{x}(t_0|t_{-1})  P_{x_i}^{-}H_0^T\left(H_0P_{x_i}^{-}H_0^T + R_y\right)^{-1}\left[h\left(\hat{x}(t_0|t_{-1})\right)-y(t_0)\right]$.

In this case, Problem \ref{eq:optim_x0} can provide a suitable value for $\hat{x}(t_0| t_{-1})$ for the new epoch
based on the last vector $\theta(t_{N-1})$ learned, that is used in eq (\ref{eq:optim_x0})and as the
initial condition $\theta(t_0| t_{-1})$ for the new epoch. We remark that when the proposed approach is run on $N_e$ epochs and $P_{x_0}^{-}$ is set equal to the value $P_{x_N}^{-}$ from the previous epoch. $x(t_0)$ is computed  next according to equation (\ref{eq:x_update}).

\section{ Training Learning Curves}
\label{app: Learning Curves}
To demonstrate the convergence of our training approach, we plot the learning curves during training of the Hodgkin Huxley, Cartpole, EMPS, Harmonic Oscillator and Yeast Glycolysis benchmarks for 20 epochs using CPU processor. 

\begin{figure}[htb]
% \hspace{0.1cm}
\centerline{\includegraphics[trim={0 0 0 0.1cm},clip,width=0.5\textwidth]{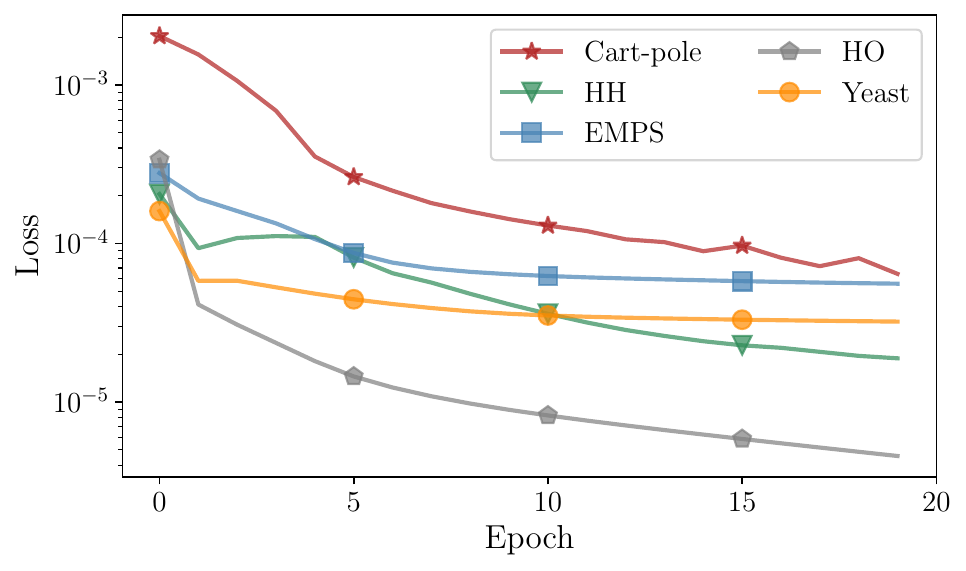}} 
\caption{ Learning curves during training of our proposed approach when benchmarked on harmonic oscillator, cart-pole, EMPS, Hodgkin-Huxley and Yeast Glycolysis dynamic models. The loss function described in eq (\ref{eq:full_optimization_function}) decreases exponentially on all benchmarks}.
\label{fig:learning_curve}
\end{figure}

results of Figure\ref{fig:learning_curve}. demonstrate the evolution of the loss function described in eq (\ref{eq:full_optimization_function}) at each epoch. The results strongly suggest that the per-epoch rate of progress of our proposed approach's loss is exponentially convergent on all of our benchmarks. 
Moreover, the results show that our approach converges in relatively very few epochs. 
The testing errors in Table \ref{tab:RMSEs}
show that our proposed method substantially outperform all of the competing methods.
This suggests that our algorithms not
only converges well, but also generalize well.

\section{Complexity analysis}\label{app: Complexity}

To calculate the complexity of our proposed training procedure, first, we need to calculate the complexity of each Jacobian. Since we are using automatic differentiation to calculate them, the complexity of getting $F_{x_{i-1}}, G_{\theta_{i-1}}$, and $F_{\theta_{i-1}}$ is $\mathcal{O}(d_x^2),  \mathcal{O}(d_{\theta}^2)$ and $\mathcal{O}(d_{\theta}d_x)$, respectively. 
However, the complexity of $G_{\theta_{i-1}}$ could be removed if $g(\theta(t_i))=0$ since $G_{\theta_{i-1}}=I$ in that case. Note that this is the case in our experiments.

Assuming $n_x \approx n_y$, the complexity of calculating  $P_{\theta_i}^{-}$,  $P_{x_i}^{-}$, $P_{x_i}$  and $P_{\theta_i}$ are $\mathcal{O}(2d_{\theta}^2d_x + d_x+ (2d_x^2 + 1)d_{\theta})$, $\mathcal{O}(2d_x^3 + d_x)$, $\mathcal{O}(4d_x^3 + 2d_x) $ and $ \mathcal{O}(d_\theta)$, respectively.

Moreover, the complexity of computing $\hat{x}(t_i)$ is $\mathcal{O}(2d_x^3+d_x^2)$, and the complexity of $\hat{\theta}(t_i)$ is $\mathcal{O}(d_{\theta}(d_x+1))$.

Since the Jacobians are differentiated once, and then evaluated at each time step, the complexity of one epoch performed over $N$ samples becomes:
\begin{equation}
    \begin{aligned}
        &\mathcal{O}(d_xd_{\theta})+ \mathcal{O}(d_{\theta}^2) + \mathcal{O}(d_{x}^2) + N[\mathcal{O}(2d_{\theta}^2d_x  +(2d_x^2+1)d_{\theta}) \\
        &+ \mathcal{O}(2d_x^3 + d_x) + \mathcal{O}(4d_x^3 + 2d_x) + \mathcal{O}(d_\theta) + \mathcal{O}(2d_x^3+d_x^2)\\ 
        &+ \mathcal{O}(d_{\theta}(d_x+1))]  
    \end{aligned}
\end{equation}
which simplifies to 
\begin{equation}
    \mathcal{O}(N(d_x^3 + d_\theta^2 d_x + d_x^2 d_\theta)).
\end{equation}
Finally, assuming $d_{\theta} \gg d_x$, the total cost of each training epoch can be simplified as:
\begin{equation}
    \begin{aligned}
        \mathcal{O}(N(2d_{\theta}^2d_x + 2d_{\theta}d_x^2))\,.
    \end{aligned}
\end{equation}
It is clear that the main source of complexity would be $d_{\theta}$.
Thus, the proposed approach may scale badly to very higher-dimensional neural network architectures. 
However, fast approximations of Newton's method exist, as pointed out by the reviewer, such as Shampoo. Although merging Shampoo with our proposed approach could reduce the computational burden, we will analyze this hypothesis in future works.
\paragraph{Test-time complexity:}
Although in our experiments we just integrated the learned mean vector field, our approach can be employed in different ways depending on the availability of data (data assimilation). The proposed approach can be also used in an online learning paradigm where learning and estimation are continuously performed. Thus, the computational complexity per time-step in these scenarios becomes $\mathcal{O}(d_\theta)$, $\mathcal{O}(d_x^3)$, and $\mathcal{O}(d_x^3 + d_\theta^2 d_x + d_x^2 d_\theta)$, for mean vector field integration, data assimilation, and continuously assimilation and adaptation, respectively.

\section{Reproducibility Checklist}

\begin{outline}
\1 This paper:

\begin{itemize}
    \item Includes a conceptual outline and/or pseudocode description of AI methods introduced (yes)
    \item Clearly delineates statements that are opinions, hypothesis, and speculation from objective facts and results (yes)
    \item Provides well marked pedagogical references for less-familiare readers to gain background necessary to replicate the paper (yes)
    \item Does this paper make theoretical contributions? (yes)
\end{itemize}

\1 In this paper:

\begin{itemize} 
    \item All assumptions and restrictions are stated clearly and formally. (yes)
    \item All novel claims are stated formally (e.g., in theorem statements). (yes)
    \item Proofs of all novel claims are included. (yes)
    \item Proof sketches or intuitions are given for complex and/or novel results. (yes)
    \item Appropriate citations to theoretical tools used are given. (yes)
    \item All theoretical claims are demonstrated empirically to hold. (yes)
    \item All experimental code used to eliminate or disprove claims is included. (no)

\end{itemize}

\1 Datasets:

\begin{itemize} 
    \item Does this paper rely on one or more datasets? (yes)
    \item A motivation is given for why the experiments are conducted on the selected datasets (no)
    \item All novel datasets introduced in this paper are included in a data appendix. (no)
    \item All novel datasets introduced in this paper will be made publicly available upon publication of the paper with a license that allows free usage for research purposes. (yes)
    \item All datasets drawn from the existing literature (potentially including authors’ own previously published work) are accompanied by appropriate citations. (yes)
    \item All datasets drawn from the existing literature (potentially including authors’ own previously published work) are publicly available. (yes)
    \item All datasets that are not publicly available are described in detail, with explanation why publicly available alternatives are not scientifically satisficing. (yes)
    
\end{itemize}

\1 Computational experiments: 
\begin{itemize} 
    \item Does this paper include computational experiments? (yes)
    \item Any code required for pre-processing data is included in the appendix. (no).
    \item All source code required for conducting and analyzing the experiments is included in a code appendix. (no)
    \item All source code required for conducting and analyzing the experiments will be made publicly available upon publication of the paper with a license that allows free usage for research purposes. (yes)
    \item All source code implementing new methods have comments detailing the implementation, with references to the paper where each step comes from (yes)
    \item If an algorithm depends on randomness, then the method used for setting seeds is described in a way sufficient to allow replication of results. (yes)
    \item This paper specifies the computing infrastructure used for running experiments (hardware and software), including GPU/CPU models; amount of memory; operating system; names and versions of relevant software libraries and frameworks. (no)
    \item This paper formally describes evaluation metrics used and explains the motivation for choosing these metrics. (partial)
    \item This paper states the number of algorithm runs used to compute each reported result. (yes)
    \item Analysis of experiments goes beyond single-dimensional summaries of performance (e.g., average; median) to include measures of variation, confidence, or other distributional information. (no)
    \item The significance of any improvement or decrease in performance is judged using appropriate statistical tests (e.g., Wilcoxon signed-rank). (yes)
    \item This paper lists all final (hyper-)parameters used for each model/algorithm in the paper’s experiments. (yes)
    \item This paper states the number and range of values tried per (hyper-) parameter during development of the paper, along with the criterion used for selecting the final parameter setting. (no)
\end{itemize}

\end{outline}

%%%%%%%%%%%%%%%%%%%%%%%%%%%%%%%%%%%%%%%%%%%%%%%%%%%%%%%%%%%%%%%%%%%%%%%%%%%%%%%
%%%%%%%%%%%%%%%%%%%%%%%%%%%%%%%%%%%%%%%%%%%%%%%%%%%%%%%%%%%%%%%%%%%%%%%%%%%%%%%

\end{document}